%% file: main.tex
\documentclass[11pt]{article}

\usepackage{algorithm}
\usepackage{algorithmic}
\usepackage{amsmath}
\usepackage{amssymb}
\usepackage{amsthm}
\usepackage{authblk}
\usepackage{booktabs}
\usepackage[margin=1in]{geometry}
\usepackage{graphicx}
\usepackage{mathtools}
\usepackage{xcolor}
\usepackage[colorlinks=true,citecolor=blue,linkcolor=blue,urlcolor=blue]{hyperref}
\usepackage[section]{placeins}

\newcommand{\R}{\mathbb{R}}
\newcommand{\E}{\mathbb{E}}
\newcommand{\N}{\mathcal{N}}
\newcommand{\F}{\mathcal{F}}

\newcommand{\MAP}{\operatorname{MAP}}
\newcommand{\Norm}[1]{\left\lVert#1\right\rVert}
\newcommand{\given}{\,\vert\,}

\title{%
  Physical-State-Guided Diffusion Sampling\\
  for Full-Waveform Inversion%
}

\author[1]{Chen Min}
\author[1]{Haowen Jiang}
\author[1,2,*]{Zheng Ma}
\author[3,*]{Xiong-Bin Yan}
\affil[1]{School of Mathematical Sciences, Shanghai Jiao Tong University, Shanghai, China}
\affil[2]{CMA-Shanghai, Shanghai Jiao Tong University, Shanghai, China}
\affil[3]{School of Mathematics and Statistics, Lanzhou University, Lanzhou, China}
\affil[*]{Corresponding authors: Zheng Ma
(\href{mailto:zhengma@sjtu.edu.cn}{zhengma@sjtu.edu.cn}); Xiong-Bin Yan
(\href{mailto:yanxb@lzu.edu.cn}{yanxb@lzu.edu.cn}).}
\date{}

\begin{document}
\maketitle

\input{content/abstract}


\input{content/introduction}
\input{content/methodology}
\input{content/psg_method}
\input{content/experiments}
\input{content/conclusion}

\section*{Acknowledgments}
Zheng Ma is supported by NSFC Grant No.~12531016 and Beijing Institute of Applied Physics and
Computational Mathematics funding HX02023-6.
Xiong-Bin Yan is supported by the Young Scientists Fund of the National Natural Science
Foundation of China (NSFC), Grant No.~12401563.
Additionally, we also thank Shanghai Institute
for Mathematics and Interdisciplinary Sciences (SIMIS) for their financial support. This research
was funded by SIMIS under grant number SIMIS-ID-2025-ST. The authors are grateful for the
resources and facilities provided by SIMIS, which were essential for the completion of this work.

\bibliographystyle{unsrt}
\bibliography{main}

\clearpage
\appendix
\input{content/appendix}

\end{document}

%% file: content/abstract.tex
\begin{abstract}
Full waveform inversion (FWI) reconstructs subsurface velocity models from
seismic measurements but faces the challenges of ill-posedness and
nonlinearity, making accurate inversion dependent on suitable initialization
and effective prior constraints.  Classical methods stabilize the inversion
and improve convergence by designing hand-crafted regularization terms, yet
these terms provide limited descriptions of complex geological structures.
Recent diffusion-based methods learn more detailed structural priors from
data and integrate them into inverse-problem solvers.
We propose Physical-State-Guided Diffusion Sampling (PSG), which decouples
diffusion posterior sampling into two mutually interacting processes:
regularized FWI optimization and conditional diffusion sampling with linear
guidance.  PSG shows overall improvements in reconstruction accuracy,
preservation of prior structure, and robustness over classical and
diffusion-based baselines across diverse OpenFWI inversion scenarios.
Through domain decomposition, PSG can generalize without retraining to more
complex, realistic velocity fields with structural patterns distinct from
those represented by the prior learned from training datasets and larger
spatial domains than the training models.  Overall, PSG offers a flexible framework for integrating learned
priors with physical constraints, with potential for extension to other
nonlinear inverse problems.
\end{abstract}

%% file: content/introduction.tex
\section{Introduction}
\label{sec:introduction}

Seismic imaging seeks quantitative descriptions of the subsurface from
wavefields recorded at or near the Earth's surface, with applications ranging
from resource exploration to environmental monitoring and seismic-hazard
assessment~\cite{yilmaz2001seismic,sambridge2002monte}.  Available techniques
include velocity analysis, traveltime tomography, migration, and linearized
inversion under the Born approximation.  Among them, full waveform inversion
(FWI)~\cite{tarantola1984inversion,virieux2009overview,fichtner2010full}
offers particularly high spatial resolution because it estimates the velocity
field by matching the phase and amplitude of simulated and observed
seismograms through a wave-equation solver.  FWI is consequently a
representative PDE-governed inverse problem in which the wave equation relates
the velocity field to seismic observations, while reconstruction remains
challenging under limited and noisy measurements.

Conventional FWI starts from an initial velocity model and iteratively
minimizes the mismatch between observed and simulated waveforms, using
gradients computed by the adjoint-state method to update the velocity through
numerical optimization~\cite{tarantola1984inversion,virieux2009overview}.
This inversion process is challenged by ill-posedness and nonlinearity.
Limited acquisition aperture and
band-limited sources leave portions of the model weakly constrained, while
measurement noise and incomplete receiver coverage further reduce the
information carried by the data~{\cite{pratt1999seismicpart2,yao2019tackling,zhang2020fwi}}.
The resulting nonconvex landscape contains many spurious basins and makes the
reconstruction strongly dependent on its initial velocity model.  This
dependence is most visible in cycle skipping, where a discrepancy of more than
approximately half a dominant period between simulated and observed arrivals
can cause a local adjoint gradient to align the wrong wave cycles and drive the
iterate toward an incorrect
model~\cite{gauthier1986two,pladys2021cycle,metivier2016measuring}.
An effective FWI method should therefore promote geologically plausible
velocity models while reducing the risk of convergence to poor local minima.

Classical approaches mitigate these difficulties through velocity
regularization~\cite{asnaashari2013regularized,esser2018total},
alternative waveform misfits~\cite{bozdaug2011misfit,warner2016adaptive},
and multiscale optimization~\cite{bunks1995multiscale}.
Tikhonov regularization penalizes spatial velocity variations to favor smooth
models, whereas total variation (TV) regularization promotes piecewise-constant
structures while allowing sharp
interfaces~\cite{asnaashari2013regularized,esser2018total,lin2014acoustic,aghamiry2018hybrid}.
These penalties constrain poorly resolved velocity components and help
stabilize reconstruction from noisy or incomplete observations.
Complementary strategies seek to reduce sensitivity to phase mismatch by
progressing from low to high frequencies or by using phase-, envelope-,
adaptive-, or optimal-transport-based
misfits~\cite{bunks1995multiscale,bozdaug2011misfit,chi2014full,warner2016adaptive,yang2018application}.
Despite these advances in stabilization and optimization, hand-crafted
penalties such as Tikhonov and TV primarily encode smoothness or gradient
sparsity, with limited ability to represent the complex spatial organization
of geological structures.

Machine learning offers additional ways to supply this structural information.
Deep learning models, primarily convolutional networks, have successfully
learned a direct mapping that approximates the inverse operator from seismic
recordings to velocity models in synthetic
settings~\cite{wu2019inversionnet,zhang2019velocitygan,zhang2020data,kazei2021mapping}.
Compared with classical iterative methods, these networks offer greater
efficiency advantages, since their inversion takes a single forward evaluation,
and they reach better accuracy on synthetic scenes~\cite{wu2019inversionnet},
where prior information learned from the training data
plays an implicit regularizing role in the ill-posed inverse
problem~\cite{arridge2019solving}.  The major limitation of
these data-driven approaches is their generalization capacity, especially under
corrupted or incomplete observations, or scenarios that do not belong to the
training data distribution~\cite{zhang2020data,antun2020instabilities,ongie2020deep}.
Physics-informed neural networks address these limitations from a different
direction, imposing the governing equation on the network through a residual
loss so that the recovered solution remains consistent with the underlying
physics~\cite{karniadakis2021physics}.  More broadly, this principle of
neural-network parameterization has been applied in geophysics both to
wavefield modeling~\cite{song2021solving} and to inversion, for which the
velocity field is parameterized by a network whose weights are updated from
observed waveforms through the wave-equation model, without requiring paired
seismic--velocity training data~\cite{yan2023unsupervised}.
Operator learning provides a framework for learning mappings
between function spaces, enabling predictions across varying physical
conditions~\cite{lu2021learning,kovachki2023neural}.
For direct FWI, recent work has demonstrated generalization
across source frequencies and locations~\cite{zhu2023fourier}, whereas taking the
forward solution operator rather than the inverse one as the learned object
yields a pretrained wavefield operator that serves as a fast differentiable
surrogate for the wave-equation solver
within an optimization-based inversion framework~\cite{yang2023rapid,huang2025physics}.  Despite these advances,
generalization remains the central limitation of these approaches, whose
accuracy degrades on velocity structures unseen during training and on data
contaminated by measurement noise, which motivates learning a more explicit and
accurate prior instead of a mapping tied to the training setting.

The rapid development of diffusion models has advanced generative modeling
and introduced a new paradigm for solving inverse problems.  By learning the
score of a family of Gaussian-corrupted marginals, these models draw samples
from the data distribution by integrating the associated reverse-time
dynamics~\cite{ho2020denoising,song2020score,song2020denoising}.
Such models can therefore be applied directly to learn the data prior of the
solution of an inverse problem, a task that requires no paired data and is
separated from the ill-posedness of the inversion
itself~\cite{daras2024survey}.  Once trained, the model supplies detailed and
expressive prior information to the inversion and accommodates variable
physical scenarios, since the observations and the differentiable forward
operator enter only at inference.  This principle has produced a broad family
of solvers for inverse problems in the image
domain~\cite{kawar2022denoising,chung2022diffusion,song2023pseudoinverse,mardani2024variational,dou2024diffusion}.
Their application to PDE-governed inverse problems, in which the forward
operator is an explicit numerical solver, remains comparatively
under-explored.  In FWI, recent work has explored several ways to couple the
learned prior with waveform data.  DiffusionFWI~\cite{wang2023prior} represents
proximal-splitting approaches, alternating waveform-consistency optimization
with reverse-sampling steps to recover velocity models consistent with the
learned prior.  Subsequent extensions to elastic FWI progressively introduce
prior information from low to high model wavenumbers, aligning the diffusion
regularization with the coarse-to-fine evolution of the
inversion~\cite{taufik2025wavenumber}.
A DAPS-style posterior sampler~\cite{taufik2025diffusion} uses clean-space
Langevin refinement and re-noising between diffusion levels.
RED-DiffEq~\cite{shan2026regularization} applies diffusion-based denoising
regularization to FWI, providing a robust numerical implementation and
achieving robust and stable FWI results on large-scale benchmarks.
Xie et al.~\cite{xie2025diffusion} introduce a score-rematching diffusion prior
directly into clean-space FWI optimization, providing a learned regularization
term for velocity reconstruction.
SGDS-FWI~\cite{shen2026diffusion} uses SDEdit~\cite{meng2022sdedit} for
initialization and then alternates waveform-consistency updates with diffusion
denoising as an approximate projection onto the learned prior manifold,
helping preserve prior consistency during optimization.
Decoupled Latent Optimization (DLO)~\cite{min2026dlo} uses a quadratic penalty
to relax the latent-optimization formulations of
DMPlug~\cite{wang2024dmplug} and D-Flow~\cite{ben2024d} into a joint optimization
over physical velocities and latent variables, allowing physically informed
initialization while the physical and latent updates constrain each other
through the coupling penalty.
Other work~\cite{peng2026robust} rescales likelihood guidance across diffusion
time and spatial locations to accommodate seismic wavefield characteristics,
stabilizing data-consistency updates.  Collectively these works demonstrate that a
learned geological prior carries structural information not explicitly encoded
by conventional smoothness or total-variation penalties, while the form in which
that prior signal is extracted from the pretrained model, and the way it is
coupled to the nonlinear wave operator,
remain open design choices.

Diffusion posterior sampling (DPS)~\cite{chung2022diffusion} is a
representative and appealing diffusion-based inverse-problem solver. It retains
conditional reverse diffusion and accommodates nonlinear differentiable forward
operators without retraining the prior, and has demonstrated its effectiveness
across a range of inverse problems.  Its practical update, however, evaluates the observation likelihood at
a single Tweedie estimate
$D_{\theta^\star}(x_t,t)\approx\mathbb E[x_0\given x_t]$ in place of the intractable noisy
likelihood obtained by integrating over $p_t(x_0\given x_t)$.  Applying this
update to FWI raises two related problems.
The first concerns the velocity model on which the wave equation is solved.
Being a conditional mean, the Tweedie estimate averages all the geological
structures that remain compatible with the current noisy state, and at high
noise levels it is therefore a blurred field rather than a velocity model that
the prior would regard as plausible.  Simulating wave propagation through such
a field yields a waveform residual, and hence a likelihood gradient, that need
not represent the residual of any realistic model; this gradient must moreover
be carried back to the diffusion state through the Jacobian of the denoiser,
so that a single scalar step size governs both how strongly the observations
are enforced and how far the state is pushed away from the prior.  Too small a
step leaves the waveforms unfitted, while too large a step drives the sample
off the prior manifold and degrades the geological structures that the diffusion
model is there to supply.
The second concerns where the trajectory begins.  Conditional sampling starts
from pure Gaussian noise, so the velocity fields denoised during the early
steps bear no direct relation to the true model, and the reverse chain offers
no natural place to insert the smoothed starting model on which conventional
FWI relies.  For an objective that is cycle
skipped once the phase misfit exceeds roughly half a dominant period, such a
starting point is unlikely to lie in the basin of attraction of the true
velocity, and the guidance returned in the early steps is correspondingly
unreliable.  Applied directly to FWI, DPS and related posterior samplers have
indeed been shown to fall short of conventional optimization supplied with a
well-chosen initial model, according to recent benchmarking
research~\cite{zheng2025inversebench}.

Here, we develop \emph{Physical-State-Guided Diffusion Sampling} (PSG), a
framework that decouples diffusion posterior sampling for FWI into two
interacting processes running over a single reverse-time trajectory.  The first
is a regularized FWI optimization over a physical velocity model, started from
an available initial velocity model as in conventional FWI and refined at every
noise level
by a waveform-fitting step that is regularized toward the current denoised
prediction of the diffusion model.  The second is a conditional sampling
process over the diffusion state, guided at each step by the updated physical
velocity through a linear identity observation model.  We model the physical
velocity as a Gaussian perturbation of a velocity drawn from the learned prior.
From the joint posterior of the clean and physical velocities, we derive
a conditional surrogate for physical refinement and an approximate
conditional score for diffusion guidance.  PSG also incorporates a domain decomposition strategy, through which a
prior trained on a synthetic benchmark can be applied directly to the inversion
of realistic velocity models that are larger than the training domain.

We demonstrate the effectiveness of PSG through extensive validation on FWI
benchmarks.  On four OpenFWI families evaluated under clean, noisy, and
missing-trace acquisitions, PSG is compared against classical regularizers and
representative diffusion-based methods, and outperforms these methods in most
of the tested settings in terms of reconstruction accuracy and preservation
of geological structure.  We
further examine the sensitivity of the reconstructions to the stochastic
sampling trajectory, and assess generalization on the Marmousi, Overthrust, and
BP2004 Salt models, which lie well outside the training distribution and are
inverted with a frozen OpenFWI prior.  These results demonstrate the accurate
and robust inversion capability of PSG as a diffusion-based solver and
suggest its potential for broader application.

\paragraph{Contributions.}
The main contributions of this work are summarized as follows.
\begin{itemize}
  \item We develop a diffusion posterior sampling scheme for FWI that runs a
  regularized FWI optimization and a conditional sampling process as two
  interacting branches of one reverse trajectory.  Beyond standard conditional
  sampling, this construction admits initialization information that need not
  lie in the noisy marginals traversed by the diffusion process, which provides
  waveform guidance from an independently initialized physical model during
  early diffusion sampling and relieves the sampling branch of the nonlinear
  task it would otherwise carry.
  The guidance it receives is instead a linear comparison of two velocity
  fields, which confines the nonlinear wave equation to the physical branch.

  \item We validate PSG on representative OpenFWI families under clean,
  Gaussian-noise, and missing-receiver conditions against direct DPS and
  established regularization and diffusion-prior methods, which demonstrates
  its accuracy on in-distribution problems and its ability to preserve sharp
  interfaces and coherent geological structures as the observations degrade.
  An additional uncertainty-quantification experiment examines reconstruction
  variability across stochastic sampling trajectories and supports the stability
  of the recovered geological structures.

  \item Through a domain decomposition strategy, we apply the OpenFWI-trained
  prior to more structurally complex, out-of-distribution velocity models with
  spatial domains larger than those used for training, namely Marmousi,
  Overthrust, and BP2004 Salt.  Together with sensitivity tests on initialization
  and observation noise, these experiments demonstrate transfer without
  retraining and suggest the potential to reconstruct complex geological
  velocity fields with varying spatial extents and structures more
  representative of real-world geology.
\end{itemize}

We first introduce FWI and its Bayesian formulation, diffusion models as data
priors, and diffusion posterior sampling in Section~\ref{sec:method}, before
presenting the proposed PSG method in Section~\ref{sec:method_psg}.
We then compare the inversion results of
PSG and baseline methods across different OpenFWI benchmark scenarios, assess
the stability of PSG reconstructions through an uncertainty quantification
experiment, and evaluate its out-of-distribution performance on the Marmousi,
Overthrust, and BP2004 Salt models in Section~\ref{sec:experiments}.
Section~\ref{sec:conclusion} summarizes the findings and future directions.
Diffusion model preliminaries and implementation details are provided in
Appendices~\ref{sec:appendix_diffusion_preliminaries}
and~\ref{sec:appendix_implementation}, respectively.

%% file: content/methodology.tex
\section{Problem Formulation and Preliminaries}
\label{sec:method}

% =============================================================================
\subsection{Full Waveform Inversion}
\label{sec:method_fwi}
% =============================================================================

Full waveform inversion (FWI) reconstructs the subsurface wave-speed field
$v(\mathbf r)$ using seismograms $d_j(t)=d(\mathbf r_j,t)$ measured at
receiver locations $\{\mathbf r_j\}_{j=1}^{N_r}$.  To this end, it adjusts
the velocity model to reduce a misfit between the recorded waveforms and
predictions obtained by solving the governing wave equation.  Because this
comparison retains information about amplitudes, phases, and multiple
arrivals, it can in principle resolve finer subsurface structures than
kinematic methods based primarily on traveltimes
~\cite{phillips1991traveltime,chen2023adjoint,hao2024topography}.
On the other hand, the strongly nonlinear dependence of the wavefield on
velocity makes the resulting optimization problem generally nonconvex.
Consequently, reliable reconstruction also depends on appropriate
regularization or prior constraints and on accounting for measurement
noise and inaccuracies in the forward model.

To formulate this inversion problem in discrete form, we first specify the acoustic
wave equation underlying the forward operator $\F$.  Under the
constant-density acoustic approximation, we consider the rectangular domain
$\Omega=[0,L_x]\times[0,L_z]\subset\mathbb R^2$ and the recording time
interval $(0,T]$.  We write $\mathbf r=(x,z)$, where $x$ denotes the
horizontal coordinate and $z$ denotes depth.  The upper edge
$\Gamma_{\mathrm{surf}}=[0,L_x]\times\{0\}$ represents the acquisition
surface.  For a source term $s(\mathbf r,t)$, the associated acoustic
pressure $u(\mathbf r,t)$ satisfies
\begin{equation}
  \label{eq:wave_equation}
  \begin{cases}
    \displaystyle
    \frac{1}{v(\mathbf r)^2}
    \frac{\partial^2u(\mathbf r,t)}{\partial t^2}
    -\nabla^2u(\mathbf r,t)
    =s(\mathbf r,t),
    & (\mathbf r,t)\in\Omega\times(0,T],\\[3pt]
    u(\mathbf r,0)=0,\quad
    \partial_t u(\mathbf r,0)=0,
    & \mathbf r\in\Omega.
  \end{cases}
\end{equation}
where $v(\mathbf r)>0$ is the unknown wave speed and $\nabla^2$ is the spatial
Laplacian.  The pressure and its time derivative are both zero at $t=0$.
We impose absorbing boundary conditions to suppress artificial reflections
caused by truncating the spatial domain.
The numerical discretization and boundary implementation are detailed in
Appendix~\ref{sec:appendix_solver}.

\begin{figure}[!htbp]
  \centering
  \includegraphics[width=0.98\linewidth]{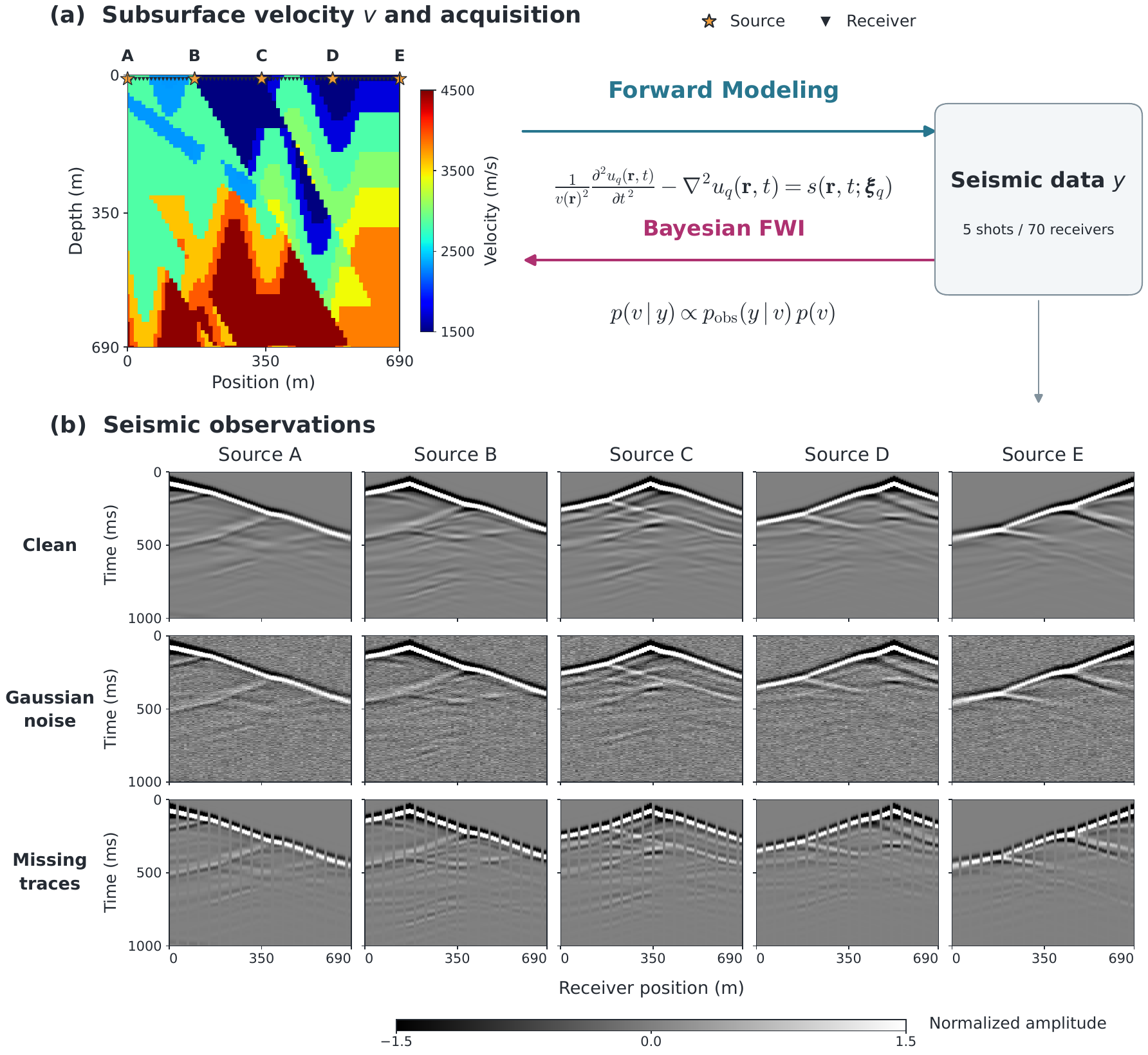}
  \caption{\textbf{Forward modeling and full waveform inversion.}
  (a) A CurveFault-B velocity model with five sources (stars) and 70 receivers
  (triangles), together with the forward mapping and inverse problem.
  The acoustic wave equation maps velocity to seismic traces, while Bayesian
  FWI combines the observation likelihood and velocity prior to form the posterior.
  (b) Corresponding shot gathers, with sources A--E arranged by column
  and observation conditions by row: clean data, additive Gaussian noise,
  and missing receiver traces.}
  \label{fig:fwi_overview}
\end{figure}

We consider a discretization of the spatial domain $\Omega$ on a
tensor-product grid with nodes $(x_i,z_j)$, where $i=1,\ldots,m_x$ and
$j=1,\ldots,m_z$.
On this grid, the wave-speed field is represented by the nodal matrix
$V=(v(x_i,z_j))\in\mathbb R^{m_x\times m_z}$.  Its vectorization, denoted by
the same symbol $v\in\mathcal X\subset\mathbb R^m$ with $m=m_xm_z$, belongs
to an admissible set $\mathcal X$ of positive velocity models.  Sources
and receivers are arranged near $\Gamma_{\mathrm{surf}}$.
For sources at $\{\boldsymbol\xi_q\}_{q=1}^{N_s}\subset\overline{\Omega}$, we solve
Eq.~\eqref{eq:wave_equation} with source term $s(\mathbf r,t;\boldsymbol\xi_q)$
and denote the resulting fields by $u_q(\mathbf r,t;v,\boldsymbol\xi_q)$.
The receiver set
$\Gamma_r=\{\mathbf r_j\}_{j=1}^{N_r}\subset\overline{\Omega}$
defines the restriction operator
\begin{equation*}
  (\mathcal P_{\mathrm{rec}} u_q)_j(t)
  :=u_q(\mathbf r_j,t;v,\boldsymbol\xi_q),
  \qquad j=1,\ldots,N_r,\quad t\in(0,T].
\end{equation*}
For each source and receiver, $(\mathcal P_{\mathrm{rec}} u_q)_j(t)$ is a continuous-time
seismic trace.  We obtain discrete recordings by sampling each trace at
times $0<t_1<\cdots<t_{N_t}\leq T$ within the recording time interval.
Collecting these sampled values across sources, sampling times, and
receivers defines the parameter-to-observation map
\begin{equation}
  \label{eq:forward_operator}
  \F:\mathcal X\longrightarrow\mathcal Y,
  \qquad
  [\F(v)]_{q,k,j}
  =(\mathcal P_{\mathrm{rec}} u_q)_j(t_k),
  \qquad
  \mathcal Y=\mathbb R^{N_s\times N_t\times N_r},
\end{equation}
which maps a velocity model to its collection of simulated shot gathers.  The
corresponding observation model is
\begin{equation}
  \label{eq:fwi_observation_model}
  y=\F(v^\dagger)+\eta,
\end{equation}
where $y\in\mathcal Y$ denotes the recorded data, $v^\dagger$ is the true
velocity, and $\eta\in\mathcal Y$ accounts for measurement noise and modeling errors.
Fig.~\ref{fig:fwi_overview} illustrates this forward-modeling and inversion setup.

Classical FWI estimates the velocity $v\in\mathcal X$ by minimizing a
waveform-misfit functional between the simulated data $\F(v)$ and the
observations $y$ with respect to $v$.  Starting from a coarse velocity
estimate, gradient-based methods update the model iteratively, typically
using the adjoint-state method to compute the misfit gradient with respect
to the velocity~\cite{tarantola1984inversion,virieux2009overview}.
Despite the high-resolution imaging enabled by full waveform information,
FWI remains a challenging problem: the nonlinear forward map and the
ill-conditioning of the inverse problem are compounded by a nonconvex
misfit landscape with multiple local minima associated with phase
ambiguity and limited illumination
~{\cite{virieux2009overview,yao2019tackling,yang2018application,pratt1999seismicpart2}}.
These difficulties become more severe when the starting velocity model
is inaccurate or the observations are affected by noise and other data
errors.
In particular, cycle skipping can occur when simulated
and observed arrivals differ by more than approximately half a dominant
period: the misfit then favors alignment of different wave cycles, which
can misdirect the velocity updates
~{\cite{pratt1999seismicpart2,pladys2021cycle,gauthier1986two}}.
These challenges motivate the use of appropriate regularization terms
or prior information to stabilize the inversion and guide the
reconstruction toward geologically plausible velocity models.

\FloatBarrier

% =============================================================================
\subsection{Bayesian Formulation of FWI}
\label{sec:method_bayesian_fwi}
% =============================================================================

Bayesian full waveform inversion treats the unknown subsurface velocity as a
random variable and seeks its posterior distribution $p(v\given y)$, which
combines information from the observations with structural knowledge about
admissible velocity models.  For fixed data $y$, agreement with the observation
model in Eq.~\eqref{eq:fwi_observation_model} enters through the likelihood.
Writing its negative logarithm as
$\ell_y(v):=-\log p_{\mathrm{lik}}(y\given v)$, up to an additive constant, gives
the general representation~\cite{dashti2013bayesian}
\begin{equation}
  \label{eq:fwi_likelihood}
  p_{\mathrm{lik}}(y\given v)\propto\exp\{-\ell_y(v)\}.
\end{equation}
A classical noise model for Eq.~\eqref{eq:fwi_observation_model} assumes
additive Gaussian error $\eta\sim\N(0,\Sigma_y)$, where $\Sigma_y$ is the
covariance in the data space and reduces to $\sigma_y^2I$ for independent and
identically distributed noise.  Under this model, the likelihood becomes
\begin{equation}
  \label{eq:gaussian_fwi_likelihood}
  p_{\mathrm{lik}}(y\given v)
  \propto
  \exp\left\{-\frac{1}{2}
  \Norm{\F(v)-y}_{\Sigma_y^{-1}}^2\right\},
  \qquad
  \Norm{r}_{\Sigma_y^{-1}}^2:=r^\top\Sigma_y^{-1}r,
\end{equation}
which corresponds to
$\ell_y(v)=\tfrac12\Norm{\F(v)-y}_{\Sigma_y^{-1}}^2$.

In FWI, limited acquisition coverage and band-limited observations can
leave parts of the velocity model weakly constrained by the data.
Fitting the observed waveforms alone may therefore yield geologically
implausible reconstructions.  Prior information provides additional
constraints that help stabilize the inversion and favor plausible
geological structures.  We incorporate this information by assigning a
prior density $p(v)$ to the velocity field.  Combining the likelihood in Eq.~\eqref{eq:fwi_likelihood} and
prior through Bayes' rule gives
\begin{equation}
  \label{eq:fwi_posterior}
  p(v\given y)
  \propto p_{\mathrm{lik}}(y\given v)p(v)
  \propto \exp\{-\ell_y(v)\}\,p(v).
\end{equation}
The resulting posterior combines the waveform consistency encoded by the
likelihood with the geological information represented by the prior,
providing a Bayesian formulation of the velocity inversion problem.

Analytic priors are commonly formed by selecting a regularity functional
$\mathcal R$ and assigning the Gibbs density
\begin{equation*}
  p(v)\propto\exp\{-\lambda\mathcal R(v)\},
  \qquad \lambda>0.
\end{equation*}
With this Gibbs prior, maximum a posteriori (MAP) estimation based on the
posterior distribution in~\eqref{eq:fwi_posterior} leads to the regularized
optimization problem~\cite{tarantola2005inverse,dashti2013bayesian}
\begin{equation}
  \label{eq:variational_fwi}
  v_{\mathrm{MAP}}
  \in\arg\min_{v\in\mathcal X}
  \bigl\{\ell_y(v)+\lambda\mathcal R(v)\bigr\},
\end{equation}
where $\mathcal X$ enforces physical admissibility and $\lambda>0$ controls
the prior strength.  Two common choices for $\mathcal R$ in Eq.~\eqref{eq:variational_fwi}
are Sobolev and total variation regularization, which impose different structural assumptions on the
velocity field.

A classical approach to encoding spatial smoothness is to assume that
the velocity field belongs to a Sobolev space $H^s(\Omega)$.  Within the
Tikhonov regularization framework, this assumption can be incorporated
by penalizing the squared Sobolev norm
~\cite{tikhonov1963regularization,tikhonov1977solutions}.  For a nonnegative
integer $s$, the corresponding regularizer is
\begin{equation}
  \label{eq:sobolev_regularizer}
  \mathcal R_{H^s}(v)
  =\Norm{v}_{H^s(\Omega)}^2
  :=\sum_{|\alpha|\leq s}\Norm{D^\alpha v}_{L^2(\Omega)}^2.
\end{equation}
Here $\alpha$ is a spatial multi-index, $|\alpha|$ denotes its total order,
and $D^\alpha v$ is the corresponding weak derivative.  The choice of the
Sobolev order $s$ in Eq.~\eqref{eq:sobolev_regularizer} allows the regularization to balance the recovery of
large-scale velocity structures with the suppression of fine-scale
oscillations.  In this work, we adopt the first-order Sobolev
seminorm as the Tikhonov penalty,
\begin{equation}
  \label{eq:classical_regularizers}
  \mathcal R_{\mathrm{Tik}}(v)
  =\lvert v\rvert_{H^1(\Omega)}^2
  =\sum_{|\alpha|=1}\Norm{D^\alpha v}_{L^2(\Omega)}^2
  =\int_\Omega\lvert\nabla v(\mathbf r)\rvert^2\,d\mathbf r.
\end{equation}
which penalizes spatial velocity variations without including a
zeroth-order term.  This quadratic penalty in Eq.~\eqref{eq:classical_regularizers} suppresses
rapid spatial variations and high-frequency artifacts, making it well suited to capturing
background velocity structures
~\cite{asnaashari2013regularized,aghamiry2018hybrid}.

Alternatively, the velocity field can be modeled in the space
$BV(\Omega)$ of integrable functions whose distributional gradients have
finite total variation~\cite{giusti1984minimal}.
This function space permits jump discontinuities and is therefore suited
to representing sharp geological interfaces.  Total variation (TV)
regularization penalizes the total variation of the distributional
gradient~\cite{rudin1992nonlinear}.  For a velocity field with an
integrable weak gradient, this penalty takes the form
\begin{equation}
  \label{eq:classical_tv_regularizer}
  \mathcal R_{\mathrm{TV}}(v)
  =\Norm{\nabla v}_{L^1(\Omega)}
  :=\int_\Omega\lvert\nabla v(\mathbf r)\rvert\,d\mathbf r,
\end{equation}
where $\lvert\nabla v\rvert$ is the Euclidean magnitude of the gradient.
By encouraging sparsity in the gradient, TV in Eq.~\eqref{eq:classical_tv_regularizer}
favors piecewise-constant
reconstructions while retaining sharp discontinuities.  This property is
particularly useful in seismic applications, where subsurface models
often contain block-like structures associated with stratified layers,
faults, and salt boundaries.  Compared with quadratic smoothing penalties,
TV better preserves discontinuities and reduces excessive smoothing of
geological interfaces, thereby favoring geologically plausible
reconstructions
~\cite{esser2018total,lin2014acoustic,aghamiry2018hybrid}.

By encoding assumptions about spatial smoothness or gradient sparsity,
these classical regularizers supplement the information provided by
waveform data.  In doing so, they help stabilize velocity reconstruction
and suppress spurious fluctuations arising from noise and incomplete data
constraints.  Nevertheless, conventional function-space regularizers
have limited capacity to capture fine-scale details and complex geological
features.  The explicit constraints they impose can further introduce
characteristic reconstruction biases: TV may represent smoothly varying
structures as staircase patterns, whereas Tikhonov regularization can blur
sharp interfaces.  These limitations have motivated the development of
data-driven regularization strategies~\cite{arridge2019solving,lunz2018adversarial,mosser2020stochastic},
among which diffusion-based prior
learning has become a leading approach for its capacity to characterize and
sample from complex prior manifolds underlying the data distribution.

% =============================================================================
\subsection{Diffusion Generative Models as Data Priors}
\label{sec:diffusion_prelim}
% =============================================================================

The forward state evolution underlying a score-based diffusion model is
specified on $t\in[0,T]$ by the It\^o stochastic differential equation
(SDE)~\cite{song2020score}
\begin{equation}
  \label{eq:forward_diffusion_sde}
  \mathrm{d}x_t
  =f(x_t,t)\,\mathrm{d}t+g(t)\,\mathrm{d}w_t,
  \qquad
  x_0\sim p_{\mathrm{data}}.
\end{equation}
Here $x_0\in\R^d$, with $d=m$ for the discretized velocity field, denotes
an unperturbed velocity model drawn from the unknown
data density $p_{\mathrm{data}}$, which represents the geological prior;
$w_t$ is a
standard $d$-dimensional Wiener process, and $f$ and $g$ are respectively the
drift and diffusion coefficients.  The coefficients are chosen so that the
geological structure present at $t=0$ is gradually obscured and the terminal
law at $t=T$ approaches a tractable reference distribution.

When the drift and diffusion coefficients satisfy the usual regularity
conditions, each state $x_t$ has a density $p_t$ governed by the
Fokker--Planck equation~\cite{song2020score}
\begin{equation*}
  \frac{\partial p_t(x)}{\partial t}
  =-\nabla_x\!\cdot\!\left(f(x,t)p_t(x)\right)
  +\frac{1}{2}g(t)^2\Delta_x p_t(x),
  \qquad p_0=p_{\mathrm{data}}.
\end{equation*}
Let $q_{0t}(x_t\given x_0)$ denote the transition density induced by
Eq.~\eqref{eq:forward_diffusion_sde}.  Marginalizing against the data density
defines the continuous family of noisy marginals
\begin{equation*}
  p_t(x_t)
  =\int_{\R^d}q_{0t}(x_t\given x_0)p_{\mathrm{data}}(x_0)\,\mathrm{d}x_0,
  \qquad t\in[0,T],
\end{equation*}
which connects the structured data distribution $p_0=p_{\mathrm{data}}$ to
the terminal reference density $p_T$ through progressively perturbed
intermediate distributions.

The variance-preserving (VP) process specializes
Eq.~\eqref{eq:forward_diffusion_sde} to the linear SDE
\begin{equation}
  \label{eq:vp_sde}
  \mathrm{d}x_t
  =-\frac{1}{2}\beta(t)x_t\,\mathrm{d}t
  +\sqrt{\beta(t)}\,\mathrm{d}w_t,
  \qquad \beta(t)>0,
\end{equation}
where $\beta(t)$ specifies the rate of corruption
~\cite{ho2020denoising,song2020score}.  Defining
$\bar\alpha_t=\exp\{-\int_0^t\beta(s)\,\mathrm{d}s\}$ gives the closed-form
Gaussian transition kernel for the process in Eq.~\eqref{eq:vp_sde}
\begin{equation}
  \label{eq:vp_transition_kernel}
  q_{0t}(x_t\given x_0)
  =\N\!\left(x_t;
    \sqrt{\bar\alpha_t}\,x_0,
    (1-\bar\alpha_t)I\right),
\end{equation}
or, equivalently, the reparameterization
\begin{equation}
  \label{eq:vp_corruption}
  x_t
  =\sqrt{\bar\alpha_t}\,x_0
  +\sqrt{1-\bar\alpha_t}\,\epsilon,
  \qquad \epsilon\sim\N(0,I).
\end{equation}
The relative noise variance
$\sigma_t^2=(1-\bar\alpha_t)/\bar\alpha_t$ increases as
$\bar\alpha_t$ decreases, whereas its reciprocal gives the signal-to-noise
ratio.  A schedule satisfying $\bar\alpha_T\approx0$ makes
$p_T\approx\N(0,I)$ and nearly independent of the initial velocity model.

Generation follows the same marginal family in the opposite direction.
For a state-independent diffusion coefficient, the time reversal of
Eq.~\eqref{eq:forward_diffusion_sde} satisfies
\begin{equation}
  \label{eq:reverse_time_sde}
  \mathrm{d}x_t
  =\left[f(x_t,t)-g(t)^2\nabla_{x_t}\log p_t(x_t)\right]\mathrm{d}t
  +g(t)\,\mathrm{d}\bar w_t,
  \qquad t:T\rightarrow0,
\end{equation}
where $\bar w_t$ is a Wiener process under the reverse-time parameterization
and $\mathrm{d}t$ is negative~\cite{song2020score}.  Starting from
$x_T\sim p_T$, this dynamics recovers $p_0$ when the marginal score
$\nabla_{x_t}\log p_t(x_t)$ is exact.  The deterministic probability-flow ODE
\begin{equation}
  \label{eq:probability_flow_ode}
  \frac{\mathrm{d}x_t}{\mathrm{d}t}
  =f(x_t,t)-\frac{1}{2}g(t)^2\nabla_{x_t}\log p_t(x_t)
\end{equation}
has the same one-time marginals and provides an alternative transport through
the distribution family~{\cite{song2020score,song2020denoising,karras2022elucidating}}.

The marginal score is unavailable directly because
$p_{\mathrm{data}}$ is represented only by samples.  The Gaussian transition
in Eq.~\eqref{eq:vp_transition_kernel}, however, permits a denoiser
$D_\theta(x_t,t)$ to be trained with the computable objective~\cite{daras2024survey}
\begin{equation}
  \label{eq:denoiser_training}
  \theta^\star
  =\arg\min_\theta
  \E_{\substack{
      t\sim\rho,\;x_0\sim p_{\mathrm{data}},\\
      \epsilon\sim\N(0,I)}}
  \left[
    \lambda(t)
    \Norm{
      D_\theta\!\left(
        \sqrt{\bar\alpha_t}x_0
        +\sqrt{1-\bar\alpha_t}\epsilon,t
      \right)-x_0
    }_2^2
  \right],
\end{equation}
where $\rho$ samples diffusion times and $\lambda(t)>0$ balances the noise
levels.  At the population optimum of the objective in Eq.~\eqref{eq:denoiser_training}
over unrestricted denoisers, the denoiser returns the conditional mean
$D_{\theta^\star}(x_t,t)=\E[x_0\given x_t]$, which determines the marginal
score through Tweedie's formula~\cite{efron2011tweedie},
\begin{equation}
  \label{eq:tweedie_score}
  \nabla_{x_t}\log p_t(x_t)
  =
  \frac{\sqrt{\bar\alpha_t}\,\E[x_0\given x_t]-x_t}{1-\bar\alpha_t}
  \mathrel{=}
  \frac{\sqrt{\bar\alpha_t}\,D_{\theta^\star}(x_t,t)-x_t}{1-\bar\alpha_t}.
\end{equation}

The density evolution underlying these dynamics, the VP--VE correspondence,
and the connection between denoising and score matching are developed in
Appendix~\ref{sec:appendix_diffusion_preliminaries}.

Training a diffusion model therefore characterizes high-quality prior
information about the solution from data alone, independently of the
specific forward operator and separated from the ill-posedness of the inverse
problem~\cite{daras2024survey}.  Embedding this
pretrained prior into the solution of an inverse problem, however, requires a
carefully designed algorithm that specifies how the prior signal is extracted
from the pretrained model and how it interacts with the physics-based data
fidelity.

% =============================================================================
% =============================================================================
\subsection{Diffusion Posterior Sampling}
\label{sec:method_dps}
% =============================================================================

Diffusion posterior sampling (DPS) modifies the reverse process of
Section~\ref{sec:diffusion_prelim} to draw samples conditioned on an
observation~\cite{chung2022diffusion}.  Given the noisy measurement
$y=\mathcal A(x_0)+\eta$ with $\eta\sim\N(0,\sigma_y^2I)$ and a forward
operator $\mathcal A$, which corresponds to the FWI forward operator
$\F$ defined in Eq.~\eqref{eq:forward_operator}, the posterior targeted by DPS is
\begin{equation}
  \label{eq:dps_target_posterior}
  \begin{aligned}
  p(x_0\given y)
  &=\frac{p_{\mathrm{prior}}(x_0)p_{\mathrm{lik}}(y\given x_0)}{p(y)}
  \propto p_{\mathrm{prior}}(x_0)p_{\mathrm{lik}}(y\given x_0),\\
  p_{\mathrm{lik}}(y\given x_0)
  &=\N\bigl(y;\mathcal A(x_0),\sigma_y^2I\bigr)
  \propto\exp\left\{-\frac{\Norm{y-\mathcal A(x_0)}_2^2}{2\sigma_y^2}\right\}.
  \end{aligned}
\end{equation}
Here $p_{\mathrm{prior}}=p_{\mathrm{data}}$ denotes the geological prior
learned by the diffusion model.  The likelihood reweights this prior,
assigning greater weight to velocity fields whose simulated waveforms
better fit the observations.  Sampling from this posterior replaces the
marginal score in Eq.~\eqref{eq:reverse_time_sde} by its conditional counterpart,
yielding the reverse-time SDE
\begin{equation*}
  \mathrm{d}x_t
  =\left[
    f(x_t,t)-g(t)^2\nabla_{x_t}\log p_t(x_t\given y)
  \right]\mathrm{d}t
  +g(t)\,\mathrm{d}\bar w_t,
  \qquad t:T\rightarrow0.
\end{equation*}
The remaining sampling procedure follows the unconditional reverse process of
Section~\ref{sec:diffusion_prelim}, so conditional generation reduces to
evaluating the conditional score, which Bayes' rule decomposes as
\begin{equation}
  \label{eq:conditional_score_split}
  \nabla_{x_t}\log p_t(x_t\given y)
  =\nabla_{x_t}\log p_t(x_t)
  +\nabla_{x_t}\log p_t(y\given x_t),
\end{equation}
where the first term is the learned prior score supplied by
Eq.~\eqref{eq:tweedie_score}, whereas the second carries the likelihood
guidance and is generally intractable because the observation depends on the
unperturbed velocity $x_0$ rather than directly on $x_t$.  Conditional
independence of $y$ and $x_t$ given $x_0$ yields
\begin{equation}
  \label{eq:exact_noisy_likelihood}
  p_t(y\given x_t)
  =\int p_{\mathrm{lik}}(y\given x_0)\,p_t(x_0\given x_t)\,\mathrm{d}x_0,
\end{equation}
where
$p_t(x_0\given x_t)=p_{\mathrm{prior}}(x_0)q_{0t}(x_t\given x_0)/p_t(x_t)$
is the normalized conditional density of the clean velocity given the
noisy state.  Equation~\eqref{eq:exact_noisy_likelihood} averages the
clean-data likelihood over this conditional distribution.
DPS replaces this distribution by its denoiser mean
$\widehat x_0(x_t)=D_{\theta^\star}(x_t,t)$ and evaluates the likelihood at that single
estimate,
\begin{equation}
  \label{eq:dps_point_approximation}
  p_t(y\given x_t)
  \approx
  p_{\mathrm{lik}}\bigl(y\given \widehat x_0(x_t)\bigr),
  \qquad \widehat x_0(x_t)=D_{\theta^\star}(x_t,t).
\end{equation}
Under the Gaussian observation model in Eq.~\eqref{eq:gaussian_fwi_likelihood}
above, the point substitution in
Eq.~\eqref{eq:dps_point_approximation} gives
\begin{equation}
  \label{eq:dps_likelihood_score}
  \nabla_{x_t}\log p_t(y\given x_t)
  \approx
  -\frac{1}{2\sigma_y^2}\nabla_{x_t}
  \Norm{y-\mathcal A(D_{\theta^\star}(x_t,t))}_2^2.
\end{equation}
Using this approximate likelihood gradient from
Eq.~\eqref{eq:dps_likelihood_score} in
Eq.~\eqref{eq:conditional_score_split} with a scalar guidance coefficient
gives the practical DPS conditional score
\begin{equation}
  \label{eq:dps_guided_score}
  \nabla_{x_t}\log p_t(x_t\given y)
  \approx
  \nabla_{x_t}\log p_t(x_t)
  -\rho\nabla_{x_t}
  \Norm{y-\mathcal A(D_{\theta^\star}(x_t,t))}_2^2,
\end{equation}
where $\rho>0$ is a scalar guidance-strength parameter that controls the
contribution of the data-consistency term~\cite{chung2022diffusion}.\footnote{%
Chung et al.~\cite{chung2022diffusion} use the residual-normalized step size
$\zeta_i=\zeta'/\Norm{y-\mathcal A(\widehat x_0(x_i))}_2$, with fixed
$\zeta'>0$, to promote stable and reliable reconstruction.}

Diffusion-based inverse-problem solvers exemplified by DPS have achieved substantial
progress in image inverse problems through likelihood approximations,
projection and splitting schemes, and annealed or variational
updates~\cite{song2023pseudoinverse,zhu2023diffpir,zhang2025daps,mardani2024variational,daras2024survey}.
Nevertheless, their direct application to FWI can be unreliable, with
benchmark results showing larger reconstruction errors and waveform misfits
than classical FWI with a suitable initialization, as well as pronounced
hyperparameter sensitivity in some methods~\cite{zheng2025inversebench}.
Several FWI-specific difficulties help
explain these limitations.

\emph{(i) Cycle skipping and nonconvexity.}
The nonlinear $\ell_2$ waveform misfit admits numerous local minima associated
with incorrect phase alignment, making FWI sensitive to the initial velocity
model.  Starting from Gaussian noise, DPS does not ensure that its early
denoised estimates provide a sufficiently accurate background velocity for
the observed data.  These estimates may therefore lie outside the basin of
attraction of the desired solution, where likelihood-based guidance can steer
the updates toward cycle-skipped local minima rather than improve the velocity
reconstruction.

\emph{(ii) Unbalanced waveform guidance.}
The quadratic residual emphasizes high-amplitude arrivals while suppressing
weaker phases that carry information about deep or poorly illuminated
regions.  Although residual normalization adapts the overall guidance
strength at each sampling step, it does not account for spatial variations
in the sensitivity of the waveform misfit to the velocity field, which can
lead to uneven updates across the model domain~\cite{peng2026robust}.

\emph{(iii) Nonphysical denoiser estimates.}
At every reverse step, DPS evaluates the forward operator on
$D_{\theta^\star}(x_t,t)$, which approximates the posterior mean
$\E[x_0\given x_t]$ rather than a sample from the learned prior.  This averaged
velocity field need not lie on the high-density prior manifold and, especially
at the high noise levels encountered early in sampling, often appears as a
blurred, nongeological intermediate image.  Because wave propagation is
nonlinear, the wavefield simulated from this averaged velocity can differ
substantially from those associated with plausible prior samples, producing
unreliable likelihood gradients.

To address these difficulties, we develop Physical-State-Guided Diffusion
Sampling (PSG), which combines conventional FWI updates with diffusion
sampling.  PSG maintains a physical velocity model initialized from a
conventional FWI starting model.  At each sampling step, it updates this
model to fit the observed waveforms, with regularization provided by the
current denoised estimate.  The updated physical model in turn serves as a
condition and provides guidance for diffusion posterior sampling.
Wave propagation
is evaluated only on the physical model, which is carried forward throughout
sampling.  We present these components in Section~\ref{sec:method_psg}.

% =============================================================================

%% file: content/psg_method.tex
\section{Physical-State-Guided Diffusion Sampling}
\label{sec:method_psg}

Recall that the DPS target in Eq.~\eqref{eq:dps_target_posterior} reweights
the geological prior by the observation likelihood.
PSG introduces an auxiliary physical velocity $\hat v=x_0+\hat\eta$,
where $\hat\eta\sim\N(0,\Sigma)$ is a Gaussian perturbation independent
of the prior velocity $x_0$.  The corresponding conditional density is
\begin{equation}
  \label{eq:bridge_general}
  p(\hat v\given x_0)=\N(\hat v;x_0,\Sigma).
\end{equation}
Here $\Sigma$ is the fixed symmetric positive-definite covariance matrix
of $\hat\eta$, describing the perturbation magnitude and correlations
across spatial locations.  In the isotropic setting, we assume
$\Sigma=\sigma_c I$, where $\sigma_c>0$ is the scalar coupling variance.

The observed waveforms are modeled as noisy measurements generated
by $\hat v$, namely $y=\F(\hat v)+\eta$ with
$\eta\sim\N(0,\sigma_y^2I)$ independent of the prior and coupling noise.
These conditional assumptions, together with the original prior
$p_{\mathrm{prior}}(x_0)$, yield the joint posterior
\begin{equation}
  \label{eq:psg_joint_posterior}
  \begin{aligned}
  \widetilde\pi(x_0,\hat v\given y)
  &=\frac{p_{\mathrm{prior}}(x_0)p(\hat v\given x_0)
       p_{\mathrm{lik}}(y\given\hat v)}{\widetilde p(y)},\\
  p_{\mathrm{lik}}(y\given\hat v)
  &=\N(y;\F(\hat v),\sigma_y^2I).
  \end{aligned}
\end{equation}
Here $\widetilde p(y)$ normalizes the joint density over $x_0$ and $\hat v$.

Marginalizing the auxiliary velocity $\hat v$ gives a posterior over
$x_0$ conditioned on the observations $y$.
As $\Sigma\to0$, the Gaussian coupling converges weakly to the Dirac
distribution at $x_0$, i.e., $\N(x_0,\Sigma)\Rightarrow\delta_{x_0}$.
Assuming that the observation likelihood $p_{\mathrm{lik}}(y\given\hat v)$
is continuous in the velocity,
this convergence gives the following limit for the marginal posterior,
with $\widetilde p(y)\to p(y)$~\cite{vono2019split}:
\begin{equation}
  \label{eq:prior_marginal_limit}
  \begin{aligned}
  \widetilde\pi(x_0\given y)
  &=\frac{p_{\mathrm{prior}}(x_0)}{\widetilde p(y)}
    \int p(\hat v\given x_0)
         p_{\mathrm{lik}}(y\given\hat v)\,\mathrm d\hat v\\
  &\xrightarrow{\Sigma\to0}
    \frac{p_{\mathrm{prior}}(x_0)}{p(y)}
    \int p_{\mathrm{lik}}(y\given\hat v)\,\delta_{x_0}(\mathrm d\hat v)\\
  &=\frac{p_{\mathrm{prior}}(x_0)p_{\mathrm{lik}}(y\given x_0)}{p(y)}
    =p(x_0\given y).
  \end{aligned}
\end{equation}
Thus, after marginalizing the physical velocity, the posterior over $x_0$
recovers the DPS target in Eq.~\eqref{eq:dps_target_posterior} as
$\Sigma\to0$.  This limiting relationship motivates a relaxed formulation
of the original posterior with a finite coupling covariance.

The marginal posterior can also be expressed as a mixture over the physical
velocity:
\[
  \widetilde\pi(x_0\given y)
  =\int\widetilde\pi(x_0\given\hat v,y)
       \widetilde\pi(\hat v\given y)\,\mathrm d\hat v.
\]
PSG couples a progressively denoised state $x_t$ with an iteratively
refined physical velocity $\hat v_t$.  At each noise level, physical
refinement is guided by the conditional MAP objective given the current
diffusion state and the observed waveforms, while the physical iterate
supplies the condition for the reverse diffusion update.  The two states
evolve together through alternating updates.
To derive these updates from the joint model, we consider the noisy state
$x_t$ generated from $x_0$ through the VP transition density
$q_{0t}(x_t\given x_0)$ in Eq.~\eqref{eq:vp_transition_kernel}.
Applying this transition to the clean variable in
Eq.~\eqref{eq:psg_joint_posterior} gives
\begin{equation}
  \label{eq:psg_noisy_joint}
  \begin{aligned}
  \widetilde\pi_t(x_t,\hat v\given y)
  &=\int q_{0t}(x_t\given x_0)
       \widetilde\pi(x_0,\hat v\given y)\,\mathrm dx_0\\
  &=\frac{p_t(x_t)p_{\mathrm{lik}}(y\given\hat v)
            B_t(\hat v\given x_t)}{\widetilde p(y)}.
  \end{aligned}
\end{equation}
Here $p_t$ is the noisy marginal of the geological prior, and
\begin{equation}
  \label{eq:coupling_average}
  B_t(\hat v\given x_t)
  :=\int p(\hat v\given x_0)p_t(x_0\given x_t)\,\mathrm dx_0,
  \qquad
  p_t(x_0\given x_t)=\frac{p_{\mathrm{prior}}(x_0)
                           q_{0t}(x_t\given x_0)}{p_t(x_t)}.
\end{equation}
The joint formulation connects waveform fitting with the geological prior
through the auxiliary physical velocity.  It motivates two complementary
processes: physical refinement regularized by the denoised estimate, and
diffusion sampling guided by the physical state.  The following subsections
develop these processes, which together constitute the PSG inversion algorithm.

\subsection{Conditional MAP Formulation and Surrogate Physical Update}
\label{sec:psg_physical_update}

Given the current diffusion state $x_t$ and the observations $y$, the
physical conditional posterior follows directly by normalizing the
joint density in Eq.~\eqref{eq:psg_noisy_joint} over $\hat v$:
\begin{equation}
  \label{eq:physical_conditional_bayes}
  \begin{aligned}
  \widetilde\pi_t(\hat v\given x_t,y)
  &=\frac{\widetilde\pi_t(x_t,\hat v\given y)}
          {\widetilde\pi_t(x_t\given y)}\\
  &=\frac{p_t(x_t)p_{\mathrm{lik}}(y\given\hat v)
            B_t(\hat v\given x_t)}
          {\widetilde\pi_t(x_t\given y)\widetilde p(y)}\\
  &=\frac{p_t(x_t)}{\widetilde\pi_t(x_t\given y)\widetilde p(y)}
     \exp\{-U_t(\hat v;x_t,y)\},
  \end{aligned}
\end{equation}
where $\widetilde\pi_t(x_t\given y)$ is the noisy posterior marginal
and the physical conditional energy is
\begin{equation}
  \label{eq:physical_conditional_energy}
  U_t(\hat v;x_t,y)=\ell_y(\hat v)-\log B_t(\hat v\given x_t).
\end{equation}
Here $\ell_y(\hat v)=-\log p_{\mathrm{lik}}(y\given\hat v)$ is the full
negative log-likelihood.  The two terms in $U_t$ account for waveform
agreement and compatibility with the clean velocities supported by $x_t$.
The prefactor in Eq.~\eqref{eq:physical_conditional_bayes} is independent
of $\hat v$.  The physical conditional MAP
estimate therefore minimizes $U_t$:
\begin{equation}
  \label{eq:physical_conditional_map}
  \hat v_t^{\MAP}(x_t,y)
  \in\arg\max_{\hat v\in\mathcal X}\widetilde\pi_t(\hat v\given x_t,y)
  =\arg\min_{\hat v\in\mathcal X}U_t(\hat v;x_t,y).
\end{equation}
To facilitate optimization of the physical conditional energy $U_t$,
Jensen's inequality gives the following upper bound:
\begin{equation}
  \label{eq:physical_expected_objective}
  \begin{aligned}
  U_t(\hat v;x_t,y)
  &=\ell_y(\hat v)
    -\log\int p(\hat v\given x_0)p_t(x_0\given x_t)\,\mathrm dx_0\\
  &\leq\ell_y(\hat v)
    -\int\log p(\hat v\given x_0)p_t(x_0\given x_t)\,\mathrm dx_0\\
  &=\underbrace{\left[
       \ell_y(\hat v)+\frac12\int\Norm{\hat v-x_0}_{\Sigma^{-1}}^2
       p_t(x_0\given x_t)\,\mathrm dx_0
     \right]}_{\displaystyle\Psi_t(\hat v;x_t,y)}
     +\kappa_\Sigma.
  \end{aligned}
\end{equation}
Here $\kappa_\Sigma=\tfrac12\log((2\pi)^m\det\Sigma)$ is the Gaussian log
normalization constant for the coupling in Eq.~\eqref{eq:bridge_general},
and the bracket defines the surrogate objective
$\Psi_t$.  This objective combines the waveform misfit with the expected
squared distance to clean velocities drawn from $p_t(x_0\given x_t)$.
Since $\kappa_\Sigma$ is independent of $\hat v$, minimizing the bound
is equivalent to minimizing $\Psi_t$.  Its quadratic coupling makes the
gradient depend on the clean conditional distribution only through its mean:
\begin{equation}
  \label{eq:physical_objective_gradient}
  \begin{aligned}
  \nabla_{\hat v}\Psi_t(\hat v;x_t,y)
  &=\nabla_{\hat v}\ell_y(\hat v)
    +\int\Sigma^{-1}(\hat v-x_0)p_t(x_0\given x_t)\,\mathrm{d}x_0\\
  &=\nabla_{\hat v}\ell_y(\hat v)
    +\Sigma^{-1}\bigl(\hat v-\mu_t(x_t)\bigr),
  \end{aligned}
\end{equation}
where $\mu_t(x_t):=\E[x_0\given x_t]$ is estimated by the pretrained
denoiser $D_{\theta^\star}(x_t,t)$.  The Gaussian coupling thus provides a
regularization term that draws the physical velocity toward the current
denoised estimate.  In the isotropic setting, $\Sigma^{-1}=\sigma_c^{-1}I$,
so the inverse variance $\lambda_c:=\sigma_c^{-1}$ acts as the coupling
weight controlling the strength of this regularization.  A single
gradient-descent step using the denoised estimate gives the physical
update at sampling time $t_n$:
\begin{equation}
  \label{eq:physical_single_step}
  \hat v_{t_{n-1}}
  =\hat v_{t_n}-\eta_v\left[
    \nabla_{\hat v}\ell_y(\hat v_{t_n})
    +\lambda_c\bigl(\hat v_{t_n}-D_{\theta^\star}(x_{t_n},t_n)\bigr)
  \right],
\end{equation}
where $\eta_v>0$ is the gradient-descent step size.
Here $\hat v_{t_n}$ denotes the physical iterate, whereas
$\hat v_t^{\MAP}$ in Eq.~\eqref{eq:physical_conditional_map} denotes an
optimizer of the conditional energy $U_t$.  The update takes one step on
$\Psi_t$ and does not require the conditional MAP problem to be solved
at each noise level.  Starting from
$\hat v_{t_N}=\hat v_{\mathrm{init}}$, each physical update continues from
the previous iterate, with regularization supplied by the current
denoised field.  Numerical details are given in
Appendix~\ref{sec:appendix_implementation}.

\subsection{Conditional Diffusion Sampling Based on Physical Velocity}
\label{sec:psg_conditional_diffusion}

Recall that, given the physical velocity $\hat v$ and the observations
$y$, the conditional prior distribution of $x_0$ follows from the joint
posterior in Eq.~\eqref{eq:psg_joint_posterior} by Bayes' rule:
\begin{equation}
  \label{eq:psg_reweighted_prior}
  \widetilde\pi(x_0\given\hat v,y)
  =\frac{\widetilde\pi(x_0,\hat v\given y)}
         {\widetilde\pi(\hat v\given y)}
  =\frac{p_{\mathrm{prior}}(x_0)p(\hat v\given x_0)}{Z(\hat v)}.
\end{equation}
Here $\widetilde\pi(\hat v\given y)$ is the marginal posterior of the
physical velocity, with
$\widetilde\pi(\hat v\given y)=p_{\mathrm{lik}}(y\given\hat v)Z(\hat v)/\widetilde p(y)$,
where $Z(\hat v)=\int p_{\mathrm{prior}}(x_0)p(\hat v\given x_0)\,\mathrm dx_0$.

Applying the VP transition to the conditional distribution in
Eq.~\eqref{eq:psg_reweighted_prior} yields the noisy density
$\widetilde\pi_t(x_t\given\hat v,y)$.  Following the reverse diffusion
construction in Section~\ref{sec:diffusion_prelim}, the corresponding
conditional score replaces the prior score in Eq.~\eqref{eq:reverse_time_sde}.
Taking the logarithmic gradient of this density with respect to $x_t$,
with $\hat v$ held fixed, and substituting the definition of $B_t$ from
Eq.~\eqref{eq:coupling_average} yields
\begin{equation}
  \label{eq:diffusion_conditional_score}
  \begin{aligned}
  \nabla_{x_t}\log\widetilde\pi_t(x_t\given\hat v,y)
  &=\nabla_{x_t}\log p_t(x_t)+\nabla_{x_t}\log B_t(\hat v\given x_t)\\
  &=\nabla_{x_t}\log p_t(x_t)
    +\nabla_{x_t}\log\int p(\hat v\given x_0)
                               p_t(x_0\given x_t)\,\mathrm dx_0.
  \end{aligned}
\end{equation}
The normalization factor $Z(\hat v)$ is independent of $x_t$ and therefore
does not contribute to the score.  Equation~\eqref{eq:diffusion_conditional_score}
thus separates the prior score from the physical guidance term
$\nabla_{x_t}\log B_t(\hat v\given x_t)$.  The coupling average $B_t$
plays the role of the noisy-state likelihood in DPS,
Eq.~\eqref{eq:exact_noisy_likelihood}, with $\hat v$ as the observation
and the identity map as the observation operator.

Following the point-estimate approximation in DPS~\cite{chung2022diffusion},
the average over $p_t(x_0\given x_t)$ is approximated by evaluating the
Gaussian coupling density at the denoised estimate
$D_{\theta^\star}(x_t,t)\approx\E[x_0\given x_t]$.
Substitution into Eq.~\eqref{eq:diffusion_conditional_score} gives
\begin{equation}
  \label{eq:diffusion_conditional_score_approx}
  \begin{aligned}
  \nabla_{x_t}\log\widetilde\pi_t(x_t\given\hat v,y)
  &\approx\nabla_{x_t}\log p_t(x_t)
     +\nabla_{x_t}\log p(\hat v\given D_{\theta^\star}(x_t,t))\\
  &=\nabla_{x_t}\log p_t(x_t)
     -\frac12\nabla_{x_t}\Norm{\hat v-D_{\theta^\star}(x_t,t)}_{\Sigma^{-1}}^2\\
  &=\nabla_{x_t}\log p_t(x_t)
     +J_{D_{\theta^\star}}(x_t,t)^\top\Sigma^{-1}
        (\hat v-D_{\theta^\star}(x_t,t)).
  \end{aligned}
\end{equation}
Here $J_{D_{\theta^\star}}(x_t,t)$ is the denoiser Jacobian with respect to $x_t$.
It maps the discrepancy between the denoised and physical velocities into
a correction of the noisy state.  This expression applies to any fixed
condition $\hat v$.
In the alternating iteration, the updated physical state
$\hat v_{t_{n-1}}$ is substituted for $\hat v$ and held fixed when computing
the score at $(x_{t_n},t_n)$.

\begin{samepage}
The approximate score in Eq.~\eqref{eq:diffusion_conditional_score_approx}
determines the correction to a discrete reverse step.  Recall the VP noise schedule $\beta(t)$ and its cumulative attenuation
$\bar\alpha_t=\exp\{-\int_0^t\beta(s)\,\mathrm ds\}$~\cite{song2020score},
as introduced in Section~\ref{sec:diffusion_prelim}.
On the reverse grid $t_N>\cdots>t_0$,
the corresponding discrete coefficients are
$\alpha_{t_n}=\bar\alpha_{t_n}/\bar\alpha_{t_{n-1}}$ and
$\beta_{t_n}=1-\alpha_{t_n}$.
Using the ancestral VP mean in Algorithm~1 of DPS~\cite{chung2022diffusion}
with a fixed reverse variance, one sampling step in the isotropic setting
takes the form
\begin{equation}
  \label{eq:diffusion_update}
  \begin{aligned}
  x'_{t_{n-1}}
  &=\frac{\sqrt{\alpha_{t_n}}(1-\bar\alpha_{t_{n-1}})}
          {1-\bar\alpha_{t_n}}\,x_{t_n}
    +\frac{\sqrt{\bar\alpha_{t_{n-1}}}\,\beta_{t_n}}
          {1-\bar\alpha_{t_n}}\,D_{\theta^\star}(x_{t_n},t_n)
    +\gamma_{t_n}z_{t_n},\\[3pt]
  x_{t_{n-1}}
  &=x'_{t_{n-1}}-\frac{\xi_{t_n}\lambda_c}{2}\nabla_{x_{t_n}}
    \Norm{\hat v_{t_{n-1}}-D_{\theta^\star}(x_{t_n},t_n)}_2^2.
  \end{aligned}
\end{equation}
\end{samepage}
The first line generates an unconditional reverse proposal, and the
second corrects it using the physical velocity $\hat v_{t_{n-1}}$ obtained
from Eq.~\eqref{eq:physical_single_step}.
The noise $z_{t_n}\sim\N(0,I)$ is drawn independently at each step.
For this ancestral form, we fix the reverse variance from the VP schedule as
$\gamma_{t_n}^2=
\beta_{t_n}(1-\bar\alpha_{t_{n-1}})/(1-\bar\alpha_{t_n})$,
the variance of the Gaussian forward bridge
$q(x_{t_{n-1}}\given x_{t_n},x_0)$, following the DDPM
parameterization~\cite{ho2020denoising}.
The effective guidance step size $\xi_{t_n}\geq0$ controls the strength
of the correction.  It absorbs implementation scaling, including the
residual normalization adopted from DPS~\cite{chung2022diffusion}.
This normalization and the remaining sampling details are described in
Appendix~\ref{sec:appendix_psg_implementation}.

Combining Eqs.~\eqref{eq:physical_single_step} and
\eqref{eq:diffusion_update} gives the alternating PSG iteration.
The iteration consists of two conditional subproblems: physical
optimization given the current diffusion state and the observations,
and guided diffusion conditioned on the updated physical velocity.
Both states are retained for the next reverse step.

The two updates involve distinct approximations: physical refinement uses
the Jensen surrogate, while diffusion guidance uses a point estimate of
the coupling integral.  The effect of the physical surrogate can be
characterized through the gap between the upper bound in
Eq.~\eqref{eq:physical_expected_objective} and the conditional energy
$U_t$ in Eq.~\eqref{eq:physical_conditional_energy}.  This gap is expressed
using the reweighted density
\[
  \rho_t^{\hat v}(x_0\given x_t)
  :=\frac{p(\hat v\given x_0)p_t(x_0\given x_t)}{B_t(\hat v\given x_t)}.
\]
Substituting this reweighted density into the difference between the
upper bound and $U_t$ yields
\begin{equation}
  \label{eq:physical_variational_gap}
  \begin{aligned}
  \Psi_t(\hat v;x_t,y)+\kappa_\Sigma-U_t(\hat v;x_t,y)
  &=\log B_t(\hat v\given x_t)
    -\int p_t(x_0\given x_t)\log p(\hat v\given x_0)\,\mathrm dx_0\\
  &=\int p_t(x_0\given x_t)
    \log\frac{B_t(\hat v\given x_t)}{p(\hat v\given x_0)}\,\mathrm dx_0\\
  &=\int p_t(x_0\given x_t)
    \log\frac{p_t(x_0\given x_t)}{\rho_t^{\hat v}(x_0\given x_t)}\,\mathrm dx_0\\
  &=\operatorname{KL}\!\left(
    p_t(\cdot\given x_t)\,\middle\|\,\rho_t^{\hat v}(\cdot\given x_t)
  \right)\geq0.
  \end{aligned}
\end{equation}
The gap in Eq.~\eqref{eq:physical_variational_gap} is therefore the
Kullback--Leibler (KL) divergence between the
original clean conditional distribution and its reweighted
counterpart~\cite{blei2017variational}.  The surrogate retains the waveform
likelihood and Gaussian coupling while making its gradient depend on the
clean conditional distribution only through the mean $\E[x_0\given x_t]$,
as shown in Eq.~\eqref{eq:physical_objective_gradient}.  Physical refinement
can therefore use the same denoised estimate as diffusion guidance,
without separately estimating the mean under the physically reweighted
distribution.  The KL identity characterizes the objective gap introduced
by this computational simplification.

PSG therefore combines surrogate physical optimization with approximate
conditional diffusion, both motivated by the same joint model.
The physical iterate is updated by descent rather than drawn from its
posterior, and the diffusion step uses the updated physical state as a
fixed condition for that step.  These conditional constructions provide
the optimization objective and sampling guidance for the alternating
iteration; they do not establish exact sampling from the marginal
posterior in Eq.~\eqref{eq:prior_marginal_limit}, the joint posterior,
or the evolving conditional distributions.

\begin{figure}[!htbp]
  \centering
  \includegraphics[width=0.98\linewidth]{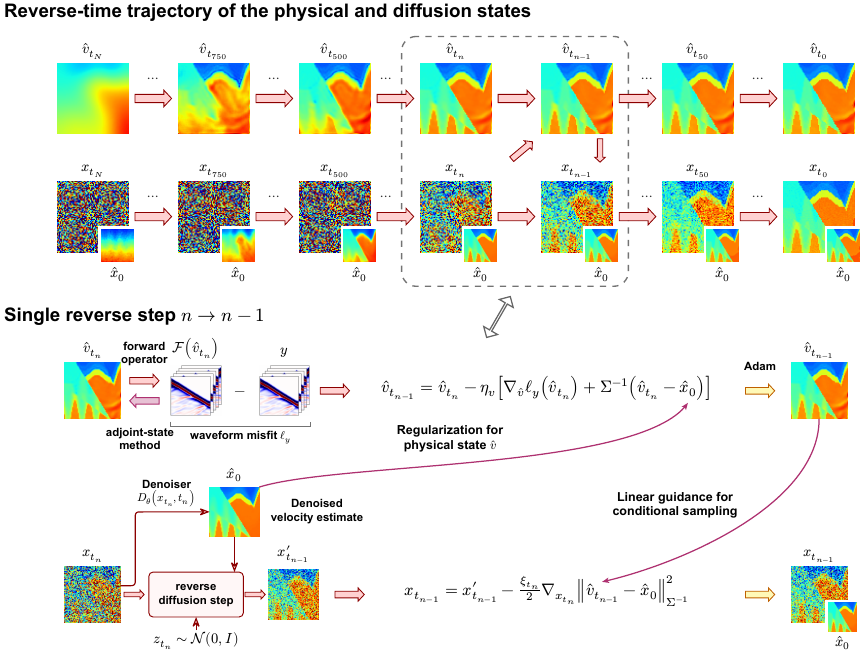}
  \caption{\textbf{Overview of Physical-State-Guided Diffusion Sampling (PSG).}
  Top: evolution of the physical velocity $\hat v_{t_n}$ and diffusion state
  $x_{t_n}$ during reverse sampling.
  Bottom: a single sampling step, comprising a regularized physical update
  and a guided diffusion update.}
  \label{fig:psg_overview}
\end{figure}

Algorithm~\ref{alg:proposed} summarizes PSG in the isotropic setting, and
Fig.~\ref{fig:psg_overview} illustrates its workflow.  Experimental details,
including hyperparameter selection, sampling discretization, and
computational cost, are provided in Appendix~\ref{sec:appendix_implementation}.
The method is stochastic through both the random initial diffusion state
and the noise increments of the reverse-time SDE.  The resulting
reconstruction variability and ensemble uncertainty are examined in
Section~\ref{sec:exp_uq}.

\FloatBarrier
\begin{algorithm}[H]
  \caption{Physical-State-Guided Diffusion Sampling (PSG)}
  \label{alg:proposed}
  \small
  \begin{algorithmic}[1]
    \REQUIRE Number of reverse steps $N$; observed waveforms $y$;
      pretrained denoiser $D_{\theta^\star}$; initial physical velocity $\hat v_{\mathrm{init}}$;
      coupling weight $\lambda_c>0$; gradient-descent step size $\eta_v$;
      reverse-time grid $\{t_n\}_{n=0}^{N}$ with $t_N>\cdots>t_0$;
      effective guidance step sizes $\{\xi_{t_n}\}_{n=1}^{N}$.
    \STATE $x_{t_N}\sim\N(0,I)$
    \STATE $\hat v_{t_N}$ is initialized as $\hat v_{\mathrm{init}}$, obtained by Gaussian smoothing of the true velocity model.
    \FOR{$n=N$ \textbf{to} $1$}
      \STATE $\hat{x}_0\leftarrow D_{\theta^\star}(x_{t_n},t_n)$
      \STATE $\hat v_{t_{n-1}}\leftarrow \hat v_{t_n}-\eta_v\left[
        \nabla_{\hat v}\ell_y(\hat v_{t_n})
        +\lambda_c\bigl(\hat v_{t_n}-\hat{x}_0\bigr)\right]$
      \STATE $z_{t_n}\sim\N(0,I)$
      \STATE $x'_{t_{n-1}}\leftarrow
        \frac{\sqrt{\alpha_{t_n}}(1-\bar\alpha_{t_{n-1}})}
             {1-\bar\alpha_{t_n}}\,x_{t_n}
        +\frac{\sqrt{\bar\alpha_{t_{n-1}}}\,\beta_{t_n}}
              {1-\bar\alpha_{t_n}}\,\hat{x}_0
        +\gamma_{t_n}z_{t_n}$
      \STATE $x_{t_{n-1}}\leftarrow x'_{t_{n-1}}
        -\frac{\xi_{t_n}\lambda_c}{2}\nabla_{x_{t_n}}
          \Norm{\hat v_{t_{n-1}}-\hat{x}_0}_2^2$
    \ENDFOR
    \RETURN $\hat v_{t_0}$ (physical output) and $\hat{x}_0$ (denoised output)
  \end{algorithmic}
\end{algorithm}

\begin{samepage}
Algorithm~\ref{alg:proposed} returns two velocity estimates: the terminal
denoised estimate $\hat{x}_0$ from the conditional sampling branch and the
final physical velocity $\hat v_{t_0}$.  In general, both results exhibit similar
overall reconstructions, with minor differences in fine-scale details.
For the main experiments on the OpenFWI benchmark in
Section~\ref{sec:exp_openfwi}, PSG takes the terminal denoised estimate
$\hat{x}_0$ as the inversion result, as it more faithfully preserves the
geological characteristics encoded by the learned prior, including sharp
boundaries, well-defined geological interfaces, and continuity within
geological regions, compared with the inversion output of the physical
optimization process.
The same output is used for the ensemble uncertainty analysis in
Section~\ref{sec:exp_uq}.
For the out-of-distribution experiments in Section~\ref{sec:exp_ood}, we take the physical
output $\hat v_{t_0}$ as the inversion result to generalize to geological structures
beyond the learned prior.  This output is not directly restricted to the
structures represented by the learned prior, allowing waveform fitting to
recover unknown physical structures, while prior-based structural
regularization promotes continuity and the recovery of geological detail.
\par
\end{samepage}

%% file: content/experiments.tex
\section{Numerical Experiments}
\label{sec:experiments}

In this section, we evaluate the inversion accuracy and stability of PSG
on established full-waveform inversion benchmarks.
Section~\ref{sec:exp_evaluation_metrics} introduces the metrics used to
assess reconstruction quality, and Section~\ref{sec:exp_openfwi} presents
the main validation results on the OpenFWI benchmark~\cite{deng2022openfwi}
across four distinct complex scenarios.
We then investigate and visualize the ensemble uncertainty of PSG across
stochastic sampling trajectories in Section~\ref{sec:exp_uq}.
Finally, Section~\ref{sec:exp_ood} evaluates the ability of this
diffusion-based inverse solver to generalize directly to
out-of-distribution velocity models, including the Marmousi, Overthrust,
and BP2004 Salt models.  Experimental details, including baseline definitions,
hyperparameter settings, sampling schemes, and computational costs,
are provided in Appendix~\ref{sec:appendix_implementation}.

\subsection{Evaluation Metrics}
\label{sec:exp_evaluation_metrics}

Reconstruction accuracy is quantified by the root mean square error
(RMSE), mean absolute error (MAE), and structural similarity index (SSIM).
Velocities are affinely normalized using a center of $3000$~m/s and a
scale of $1500$~m/s, as in Appendix~\ref{sec:appendix_pretraining}.
Let $\hat v,v\in\mathbb R^m$ denote the normalized reconstructed and
ground-truth velocity fields, respectively, on the spatial grid.
RMSE and MAE quantify pointwise deviations and are given by
\begin{equation}
  \label{eq:exp_mae_rmse}
  \mathrm{RMSE}
  =\sqrt{\frac{1}{m}\sum_{i=1}^{m}(v_i-\hat v_i)^2},
  \qquad
  \mathrm{MAE}
  =\frac{1}{m}\sum_{i=1}^{m}\bigl|v_i-\hat v_i\bigr|,
\end{equation}
with RMSE placing greater emphasis on large residuals and MAE measuring
the mean absolute deviation.  These errors in Eq.~\eqref{eq:exp_mae_rmse} are evaluated directly on the
normalized fields and multiplied by $1500$~m/s for reporting in physical
units; lower values indicate better accuracy.

For SSIM~\cite{wang2004image}, both normalized fields are clipped to
$[-1,1]$ and then linearly rescaled to $[0,1]$.  Local structural
agreement is quantified by
\begin{equation}
  \label{eq:exp_ssim}
  \mathrm{SSIM}(v,\hat v)
  =\frac{(2\mu_v\mu_{\hat v}+C_1)
          (2\sigma_{v,\hat v}+C_2)}
         {(\mu_v^2+\mu_{\hat v}^2+C_1)
          (\sigma_v^2+\sigma_{\hat v}^2+C_2)},
\end{equation}
where $\mu_v,\mu_{\hat v}$ are local averages of the rescaled
fields, $\sigma_v^2,\sigma_{\hat v}^2$ are their local variances, and
$\sigma_{v,\hat v}$ is their local covariance.  These statistics in Eq.~\eqref{eq:exp_ssim} use a
normalized $11\times11$ Gaussian window with standard deviation $1.5$
grid cells and five-cell zero padding at the grid boundaries.  We set
$C_1=0.01^2$ and $C_2=0.03^2$ for the unit data range and average the
local scores over all original grid points to obtain the reported SSIM;
higher values indicate greater structural agreement.  RMSE and MAE
characterize overall velocity-amplitude errors, whereas SSIM measures
perceptual and structural agreement, together offering complementary
assessments of inversion quality.

\FloatBarrier
\subsection{Validation on OpenFWI Datasets}
\label{sec:exp_openfwi}
\label{sec:exp_overview}

OpenFWI~\cite{deng2022openfwi} is a large-scale benchmark for full-waveform
inversion that pairs subsurface velocity models with simulated seismic
recordings across diverse synthetic geological structures.
We consider four representative
families covering distinct complex scenarios:
FlatVel-B (FV-B), with horizontal layers separated by sharp velocity
contrasts;
FlatFault-B (FF-B), with layered backgrounds displaced by faults;
CurveVel-B (CV-B), with curved and dipping interfaces exhibiting lateral
variations;
and CurveFault-B (CF-B), with curved stratigraphy intersected by faults.

We compare PSG with the following baselines: standard FWI without regularization,
Tikhonov-FWI~\cite{asnaashari2013regularized},
TV-FWI~\cite{esser2018total}, DPS~\cite{chung2022diffusion},
RED-DiffEq~\cite{shan2026regularization},
DiffusionFWI~\cite{wang2023prior}, and SGDS-FWI~\cite{shen2026diffusion}.
To ensure a fair comparison, all methods are evaluated under consistent
physical conditions, and all diffusion-based methods use the same
pretrained diffusion model for each dataset family.

\begin{figure}[!htbp]
  \centering
  \includegraphics[width=0.98\linewidth]{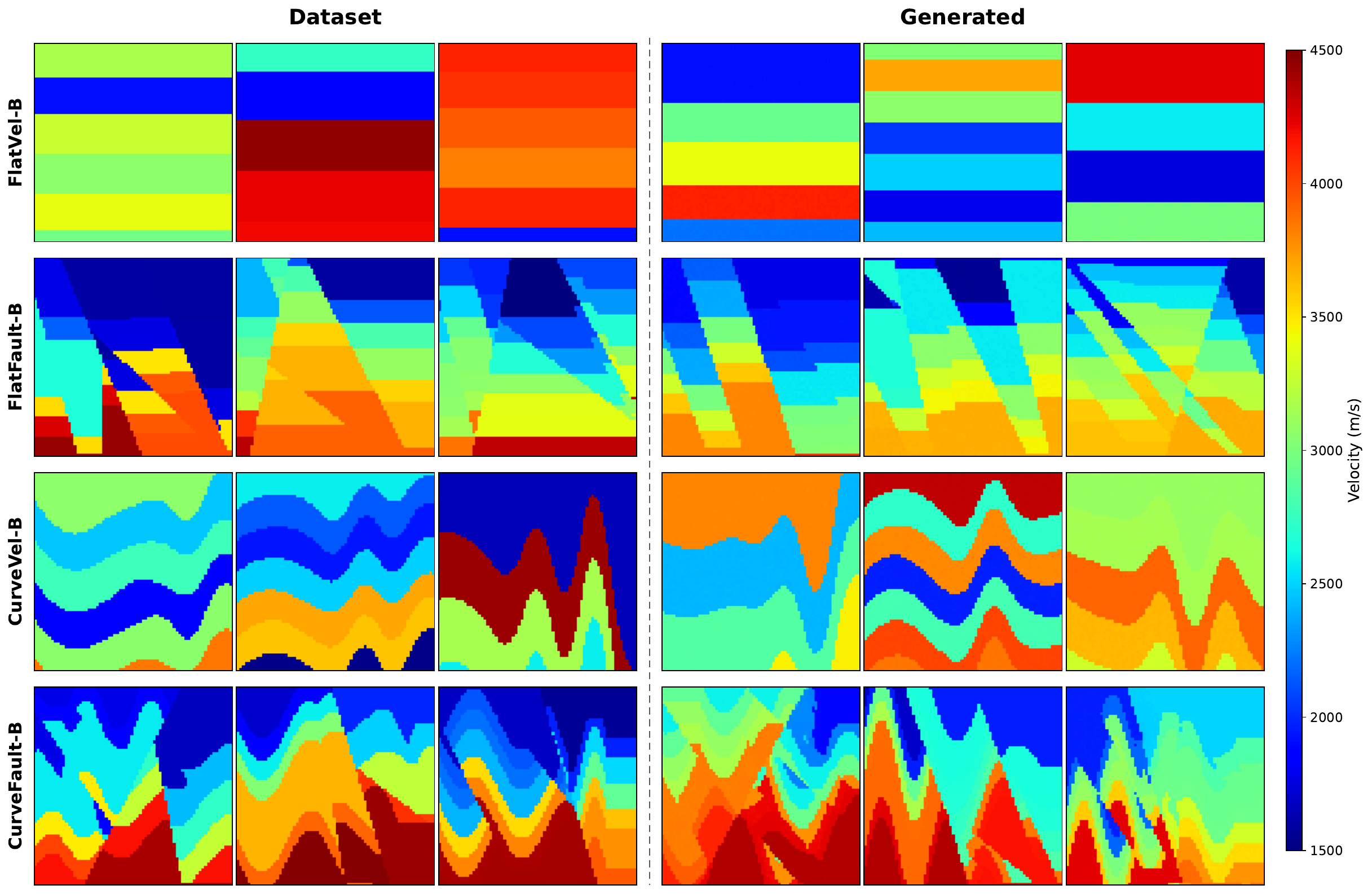}
  \caption{\textbf{Comparison of dataset samples and generated samples.}
  Left: OpenFWI test-set velocity models not used for training;
  right: unconditional samples from
  the corresponding pretrained EDM models.}
  \label{fig:datasets_vs_generated}
\end{figure}

For each family, we train a separate EDM diffusion
model~\cite{karras2022elucidating} and share the pretrained
model among all diffusion-based methods evaluated on that family.
Pretraining settings are provided in Appendix~\ref{sec:appendix_pretraining}.
This variance-exploding (VE) parameterization uses denoising training over
continuous noise levels, which naturally accommodates custom noise schedules
on a continuous time axis.  Therefore, all compared methods originally based on the
DDPM formulation~\cite{ho2020denoising} are adapted through the
time-dependent linear rescaling between VE and variance-preserving (VP)
states and the associated noise-level conversion, as in
InverseBench~\cite{zheng2025inversebench}.
Method implementation protocols and hyperparameter settings are provided
in Appendices~\ref{sec:appendix_baselines} and~\ref{sec:exp_hyperparameters}.
Fig.~\ref{fig:datasets_vs_generated} visualizes ground-truth velocity models
that were not used to train the diffusion models, alongside unconditional samples generated by
1000-step deterministic DDIM sampling~\cite{song2020denoising} of each
pretrained model,
illustrating the characteristic layered and faulted structures captured
by the learned priors.

\FloatBarrier
\subsubsection{Clean Seismic Data}
\label{sec:exp_clean_results}

We evaluate all methods on $100$ velocity--waveform pairs from the testing
dataset of each velocity family.  The observed wavefield is sampled at
$70$ receivers along the surface over $1000$ time steps.
For methods requiring an initial velocity model, we smooth the ground-truth
velocity with a Gaussian filter of standard deviation
$\sigma_{\mathrm{init}}=10$, providing the same initialization for these methods.

For optimization-based methods, including FWI, its regularized variants, and
RED-DiffEq~\cite{shan2026regularization}, we run $300$ iterations following
the benchmark configuration in the official RED-DiffEq repository.
For DiffusionFWI~\cite{wang2023prior}, we follow the inner-loop setting
reported in the original paper, using $10$ FWI optimization steps between
consecutive diffusion updates.  We use $100$ reverse diffusion steps in
our experiments.
For SGDS-FWI~\cite{shen2026diffusion}, we adapt the authors' official
implementation to the shared Adam/EDM framework, using a total of $1050$
FWI optimization steps.
DPS and PSG both use a $1000$-step discretization of the VP noise schedule
associated with DDPM~\cite{ho2020denoising}.
A per-step runtime breakdown for PSG is provided in
Appendix~\ref{sec:appendix_cost}.
Detailed algorithmic descriptions and hyperparameter settings are provided
in Appendices~\ref{sec:exp_baselines} and~\ref{sec:exp_hyperparameters};
the acoustic forward operator and its numerical discretization are detailed
in Appendix~\ref{sec:exp_protocol}.

Fig.~\ref{fig:openfwi_clean} compares the reconstructed velocity fields
for one fixed test sample from each family, alongside the ground truth
and the smoothed initial model.  Table~\ref{tab:openfwi_clean_metrics}
reports MAE, RMSE, and SSIM averaged over all $100$ test samples per family.

Unregularized FWI exhibits oscillatory artifacts and inaccurate velocity
amplitudes, particularly in deeper regions.  Tikhonov and TV regularization
reduce these artifacts, but Tikhonov tends to blur interfaces, whereas TV
preserves sharper contrasts while leaving fine structures unresolved.
Direct DPS produces velocity fields that retain the structural patterns of
the learned prior, but with misplaced interfaces and incorrect velocity
amplitudes, resulting in substantially larger errors.  This can be
attributed in part to the lack of a suitable physical initialization to
support effective gradient guidance during early sampling.
DiffusionFWI recovers the dominant layered structures, but its reconstruction
of finer details remains less accurate in the curved and faulted families.
RED-DiffEq suppresses artifacts and improves agreement with the ground truth,
achieving the strongest baseline metrics on FF-B, CV-B, and CF-B.
SGDS-FWI produces relatively sharp boundaries and clear geological patterns,
but resolves internal details near the bottom of the model less accurately.
The outputs of PSG's guided sampling process better preserve the geological
features captured by the learned prior, with sharp interfaces and coherent
fine-scale structures, including in deeper regions.  Consistent with these
visual observations, Table~\ref{tab:openfwi_clean_metrics} shows that PSG
achieves the lowest mean MAE and RMSE and the highest mean SSIM on all four
datasets, providing more accurate velocity reconstructions.

\begin{figure}[!htbp]
  \centering
  \includegraphics[width=\linewidth]{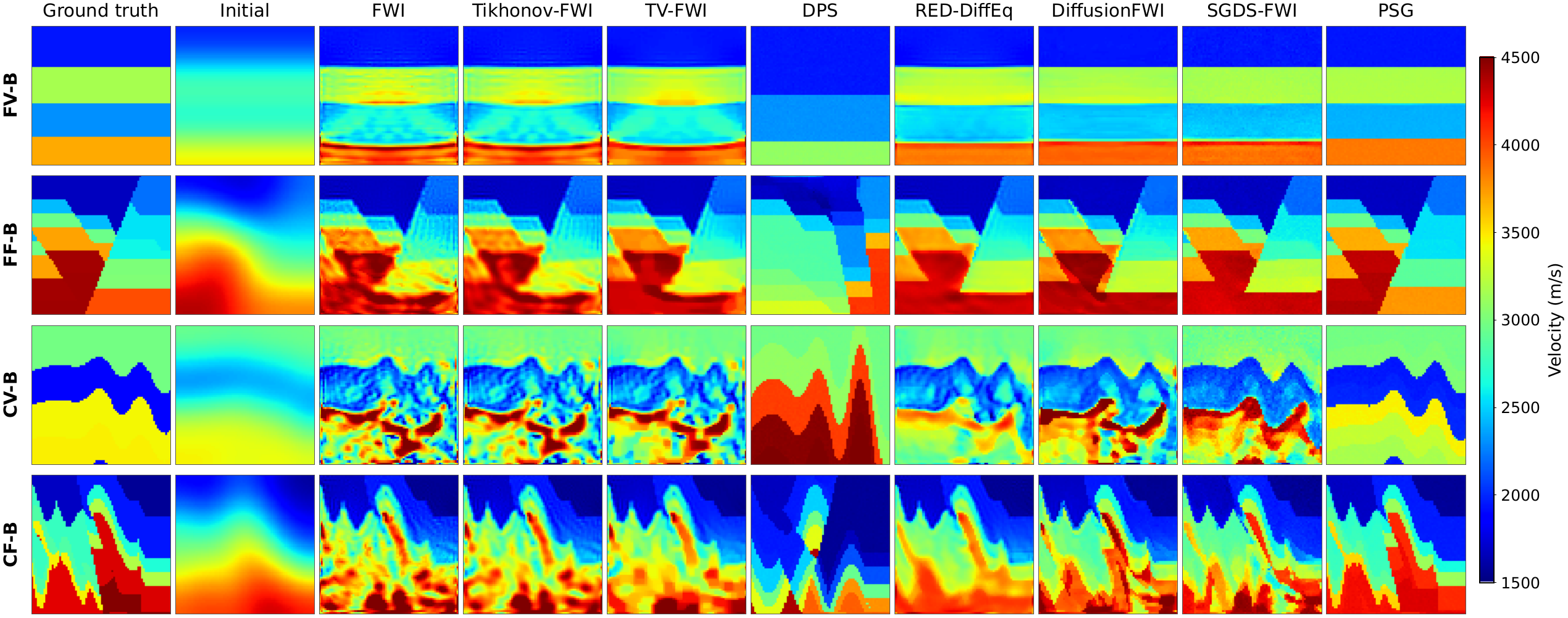}
  \caption{\textbf{OpenFWI clean-data qualitative comparison.}
  Rows show test samples from FV-B, FF-B, CV-B, and CF-B.
  Columns show the ground truth, the smoothed initial
  model, and the reconstructions from the eight methods.  All panels
  share the same velocity scale in m/s.}
  \label{fig:openfwi_clean}
\end{figure}

\input{tables/openfwi_clean_metrics}

\FloatBarrier
\subsubsection{Noisy Seismic Data}
\label{sec:exp_noise_results}
% =============================================================================

We next evaluate robustness to additive measurement noise.  The seismic
observations are perturbed by independent Gaussian noise with standard
deviations $\sigma\in\{0.1,0.3,0.5\}$ in the original amplitude scale.
These levels correspond to mean per-sample signal-to-noise ratios of
$24.1$, $14.6$, and $10.2$~dB over the 400 test models.
All methods receive the same noisy observations and retain the
initialization and hyperparameter settings of the clean-data experiment.

\begin{figure}[!htbp]
  \centering
  \includegraphics[width=\linewidth]{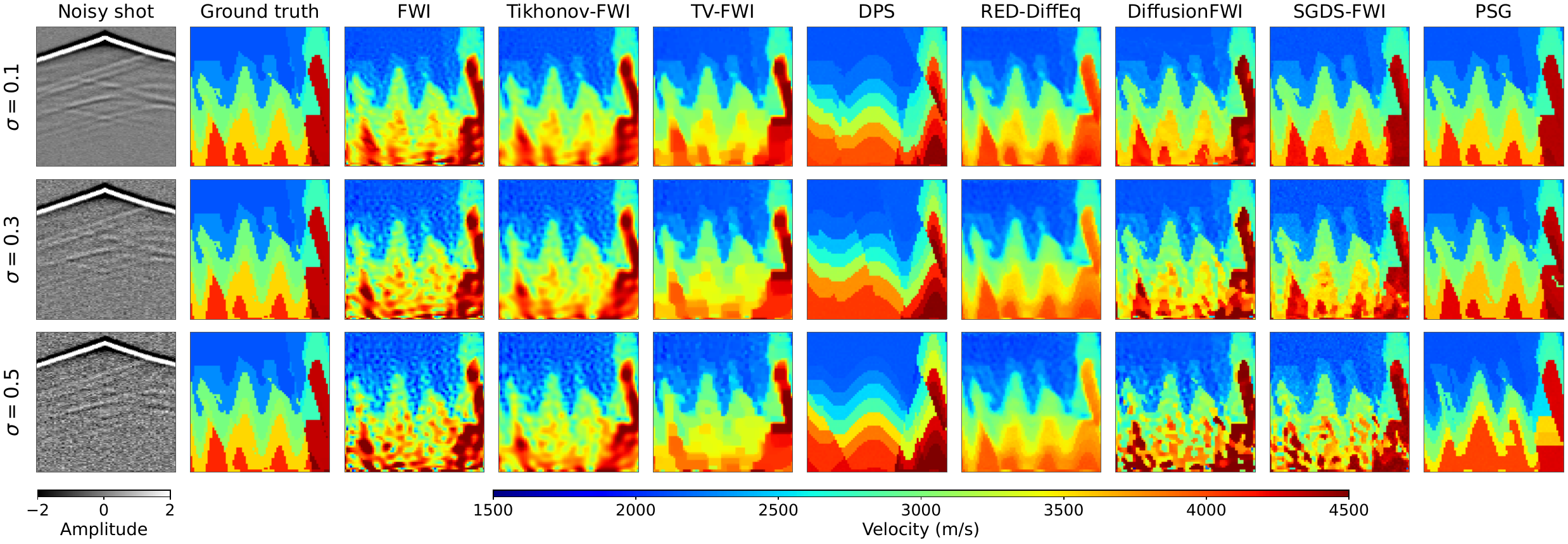}
  \caption{\textbf{OpenFWI noisy-data results (Gaussian noise).}
  Single-sample reconstructions on CF-B at three noise levels
  ($\sigma=0.1,0.3,0.5$). From left to right: the noisy shot gather,
  the ground-truth velocity model, and reconstructions from the compared methods.}
  \label{fig:openfwi_noise}
\end{figure}

\enlargethispage{\baselineskip}
Fig.~\ref{fig:openfwi_noise} shows the noisy shot gathers and
reconstructions for a sample from the CurveFault-B test set at the three
noise levels, illustrating how the reconstructions degrade as noise increases.
Table~\ref{tab:openfwi_noise_metrics} reports MAE, RMSE, and SSIM averaged
over 100 test samples from each family.

Unregularized FWI and Tikhonov-FWI develop increasingly pronounced
oscillatory artifacts as the noise level rises, particularly in deeper
regions.  TV-FWI suppresses much of this contamination, but its
piecewise-constant reconstructions introduce staircase patterns and leave
fine interfaces unresolved.  DPS continues to exhibit the behavior observed
under clean data: its outputs retain structural patterns of the learned
prior, yet misplaced interfaces and incorrect velocity amplitudes lead to
inaccurate reconstructions.  DiffusionFWI and SGDS-FWI show similar
sensitivity to noise.  Both recover recognizable structures at low noise,
but stronger perturbations introduce extensive artifacts and disrupt the
coherent layer and fault patterns of the velocity models.

PSG also exhibits substantial robustness to measurement noise, attaining
the lowest mean MAE and RMSE and the highest mean SSIM on all four families
at $\sigma=0.1$.  As noise increases, RED-DiffEq provides the most
consistent reconstruction accuracy among the diffusion baselines.
At $\sigma=0.3$, PSG attains the lowest mean MAE and highest mean SSIM
on all four families, while RED-DiffEq achieves lower RMSE on three of them.
At $\sigma=0.5$, PSG retains the lowest mean MAE on FV-B and CV-B,
although RED-DiffEq achieves lower RMSE on all four families.
Across the tested noise levels, PSG generally achieves lower errors than
DPS, DiffusionFWI, and SGDS-FWI.

As illustrated in Fig.~\ref{fig:openfwi_noise}, the degradation of PSG is
most apparent in deeper regions: the reconstructed models retain plausible
structures under the learned prior, but layer boundaries and fault geometry
deviate from the ground truth.  Despite these distortions, PSG attains the
highest mean SSIM in all 12 family--noise conditions.
This structural advantage is consistent with the distinction between
conditional sampling and diffusion-regularized physical optimization:
PSG couples a conditional diffusion trajectory to the evolving physical
state, preserving coherent prior structures even when the recovered
velocity amplitudes become less accurate.
These results support the robustness of PSG as a conditional diffusion
solver for FWI under noisy observations.

% Keep Fig. 5 with its discussion; restore normal float spacing afterwards.
\setlength{\intextsep}{4pt}
\setlength{\intextsep}{12pt plus 2pt minus 2pt}

\FloatBarrier
\input{tables/openfwi_noise_metrics}

\FloatBarrier
% =============================================================================
\subsubsection{Seismic Data with Missing Traces}
\label{sec:exp_missing_results}
% =============================================================================

We finally evaluate the methods under incomplete acquisition, with $10$,
$35$, and $60$ of the $70$ receiver traces removed from each shot gather.
These settings correspond to missing fractions of $14.3\%$, $50.0\%$,
and $85.7\%$, respectively.  For each test model, the randomly selected
missing receiver locations are shared across all shots and methods, and
the missing sets are nested across the three levels.
All methods evaluate the waveform misfit only at the retained receivers.
The initialization and hyperparameter settings follow the clean-data
experiment, and no additive noise is introduced.

\begin{figure}[!htbp]
  \centering
  \includegraphics[width=\linewidth]{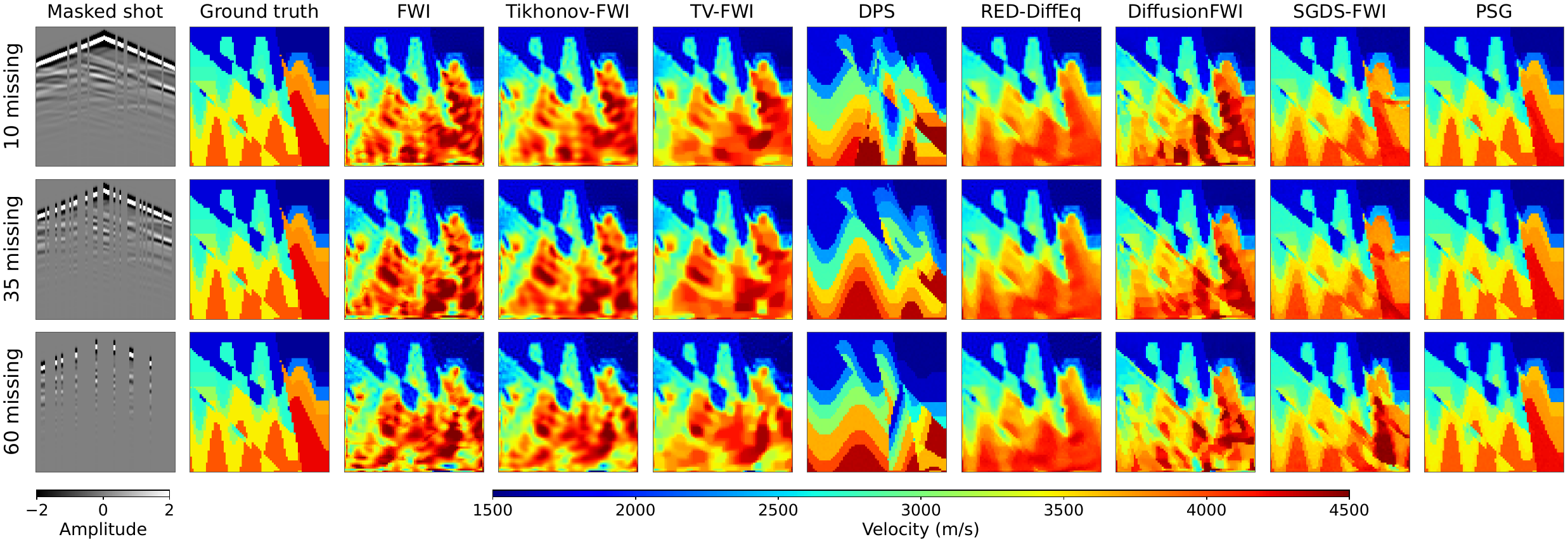}
  \caption{\textbf{OpenFWI missing-trace results.}
  Single-sample reconstructions on CF-B with $10$, $35$, and $60$ of the
  $70$ receiver traces removed. From left to right: the masked shot gather,
  the ground-truth velocity model, and reconstructions from the compared methods.}
  \label{fig:openfwi_missing}
\end{figure}

The inversion results of all methods are visualized in
Fig.~\ref{fig:openfwi_missing}.
Table~\ref{tab:openfwi_missing_metrics} reports MAE, RMSE, and SSIM
averaged over 100 test samples from each family.

Reconstruction quality generally deteriorates as the number of missing
traces increases, with the strongest degradation when only $10$ receivers
remain.  Unregularized FWI and its classical regularized variants exhibit
pronounced artifacts and a loss of fine-scale detail, particularly in
deeper regions.  DiffusionFWI and SGDS-FWI recover the dominant layer and
fault patterns, but local artifacts and inaccurate deep interfaces persist
under sparse acquisition.  DPS retains the structural character of the
learned prior but continues to produce inaccurate layer boundaries and
fault geometry.  RED-DiffEq recovers the main layer and fault patterns,
although deeper interfaces remain blurred.  PSG retains sharper interfaces and
fine-scale structural details even when $60$ of the $70$ receivers are
removed, achieving the lowest mean MAE and RMSE and the highest mean SSIM
on all four families at every missing-trace level.
Together with the Gaussian-noise results in
Section~\ref{sec:exp_noise_results}, these findings support the robustness
of PSG to both measurement noise and severely incomplete acquisition.

\FloatBarrier
\input{tables/openfwi_missing_metrics}

\FloatBarrier
% =============================================================================
\subsection{Stochastic-Trajectory Sensitivity and Ensemble Uncertainty}
\label{sec:exp_uq}

PSG is stochastic through both the initial diffusion state and the noise
increments of the reverse-time SDE.  To assess the effect of this
randomness, we select ten test models from each OpenFWI family
and perform $20$ independent runs per model,
giving $800$ inversions in total.  Each run uses an independent draw of
$x_{t_N}\sim\N(0,I)$ and of the reverse-SDE noise increments, while the
clean observations, physical initialization, pretrained prior, and
hyperparameters remain fixed as in Section~\ref{sec:exp_clean_results}.
We use the pixelwise mean of the $20$ terminal denoised velocity fields
as the ensemble prediction and their standard deviation as a measure of
variability across stochastic trajectories.

\begin{figure}[!htbp]
  \centering
  \includegraphics[width=\linewidth]{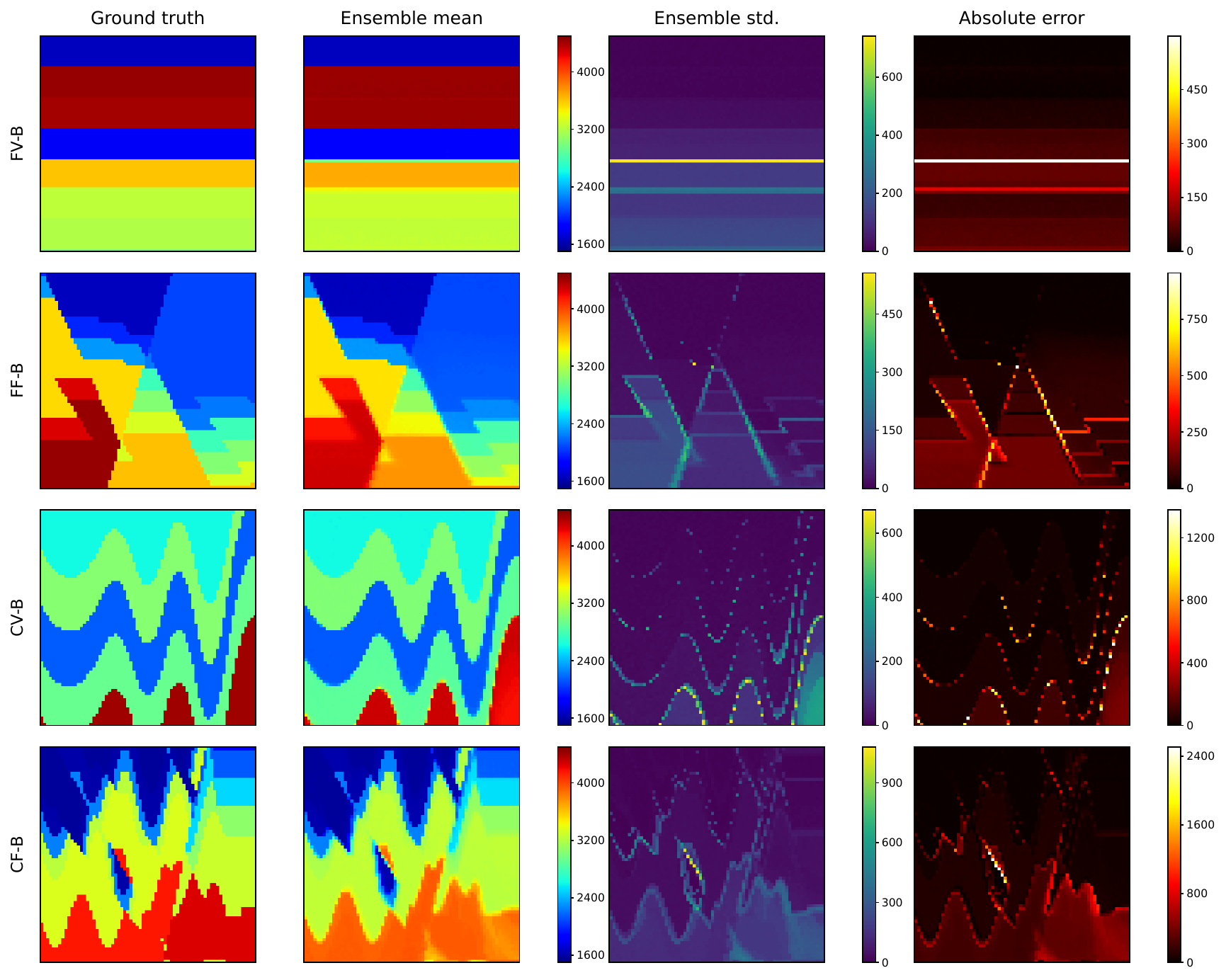}
  \caption{\textbf{Sensitivity of PSG to stochastic sampling trajectories.}
  Rows show test samples from FV-B, FF-B, CV-B, and CF-B.
  Columns show the ground truth, the ensemble mean of the denoised outputs
  over $20$ runs,
  the pixelwise ensemble standard deviation, and the absolute error
  of the ensemble mean. All color bars are in m/s; ground truth and mean
  share a velocity scale, while standard deviation and error use
  panel-specific scales.}
  \label{fig:openfwi_uq_ensembles}
\end{figure}

Fig.~\ref{fig:openfwi_uq_ensembles} shows that the ensemble means
remain close to the ground-truth velocity fields, preserving bulk velocity
contrasts, layer geometry, and fault patterns.  Variability is concentrated
primarily near faults and geological interfaces, with some additional
dispersion at depth.  The main velocity structures are comparatively stable,
while uncertainty in local interface positions and fine-scale details
spatially coincides with larger reconstruction errors.

To quantify this spatial correspondence, we compute the Spearman and
Pearson correlations between the pixelwise ensemble standard deviation
and the absolute error of the ensemble mean separately for each of the
$40$ test models.  The mean coefficients are $0.7063$ and $0.6457$,
respectively, with positive correlations in $36$ and $37$ of the $40$
models.  These results support an association between sampling variability and
local inversion error across most test models.

To visualize this relationship, we divide each model's pixels into ten
quantile groups of ensemble standard deviation, from the lowest uncertainty
(group $1$) to the highest (group $10$), and compute the mean absolute error
of the ensemble prediction within each group.  The groups have approximately
equal pixel counts and express relative uncertainty within each model.

\begin{figure}[!htbp]
  \centering
  \includegraphics[width=\linewidth]{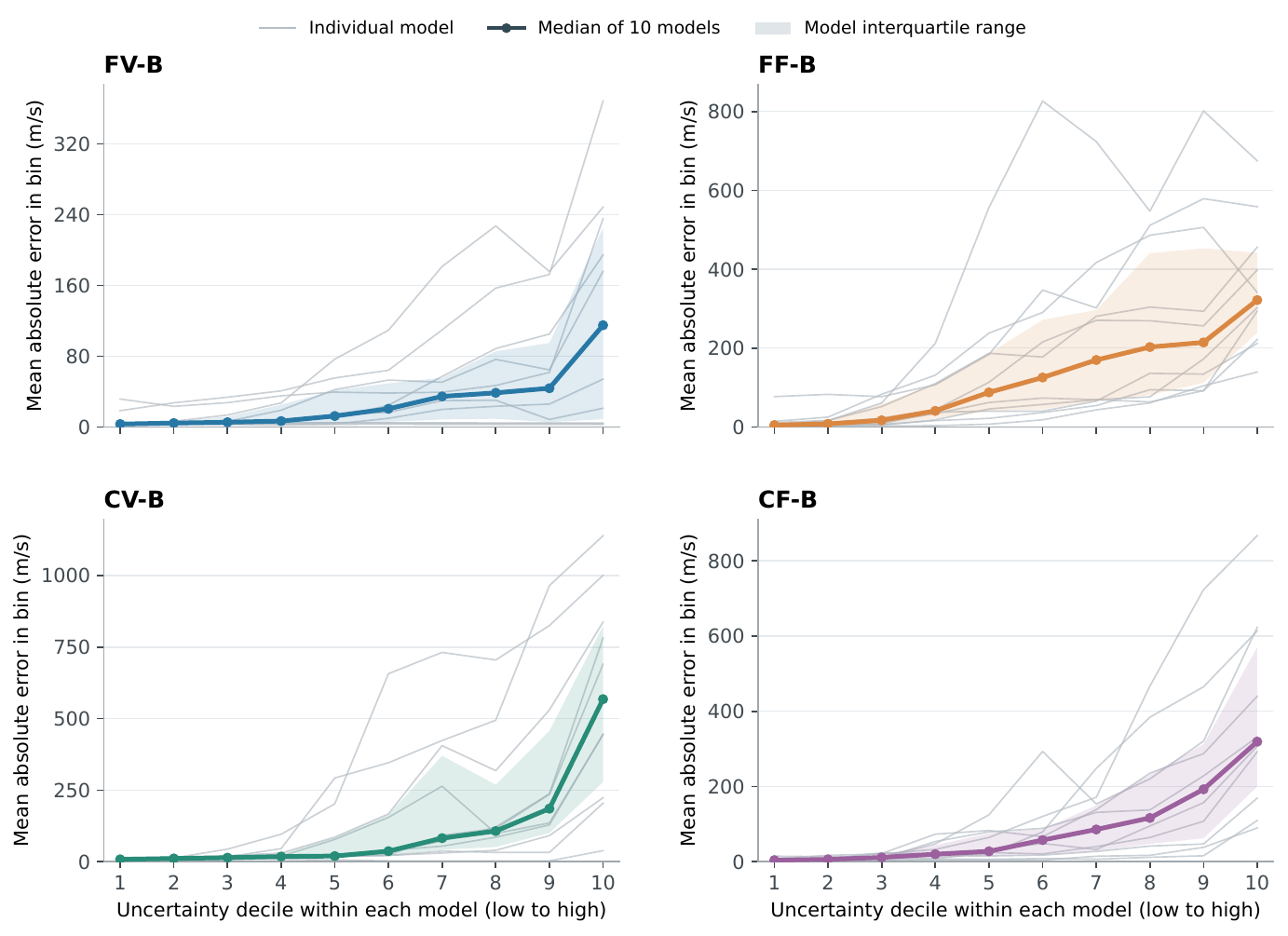}
  \caption{\textbf{Local reconstruction error across uncertainty deciles.}
  Pixels are grouped by ensemble standard deviation within each test model.
  Gray lines show the mean absolute error of the ensemble mean (m/s) in each
  group for individual models; colored lines show the median of these errors
  across ten models per family. Shading
  spans the corresponding $25$th--$75$th percentiles across models.
  Higher deciles indicate greater uncertainty; vertical scales differ
  across families.}
  \label{fig:openfwi_uq_deciles}
\end{figure}

Fig.~\ref{fig:openfwi_uq_deciles} shows that groups with higher uncertainty
generally have larger local errors across all four families.  Individual
curves retain model-specific differences and need not increase at every
decile.  This trend supports the use of ensemble dispersion to identify
relatively less accurate regions within a reconstruction.

Across the $40$ models, the mean within-model standard deviations of
RMSE, MAE, and SSIM are $40.9$~m/s, $27.5$~m/s, and $0.0266$,
respectively.  These results indicate limited sensitivity to sampling
randomness in the recovery of dominant geological structures under the
tested conditions.  Different random seeds generally preserve the main
layer and fault patterns, while local interface positions and velocity
values vary.  Ensemble dispersion provides an empirical indicator of
regions with greater local instability and typically larger reconstruction
errors.

\FloatBarrier

\subsection{Validation on Out-of-Distribution Models}
\label{sec:exp_ood}

We examine whether PSG can reconstruct velocity structures beyond the
OpenFWI training distribution using
Marmousi~\cite{versteeg1994marmousi},
Overthrust~\cite{aminzadeh19973}, and
a Salt model derived from the BP2004 velocity
benchmark~\cite{billette2005bp}.
Marmousi combines curved, dipping layers with intersecting faults and
fine-scale deep interfaces.  Overthrust contains thrust-related layer
displacements and pronounced velocity contrasts.  The Salt target is
derived from a cropped and resampled portion of the left salt body in
the BP2004 velocity benchmark.  Its steep, irregular salt boundaries
and strong velocity contrast with the surrounding medium present an
additional challenge to recovery from a smoothed initial model.
These structures test whether a prior learned from OpenFWI can support
inversion on more complex targets without retraining.

Each benchmark velocity field is resampled using area interpolation onto
a computational grid with $70$ points in depth and $190$ points horizontally.
The grid spacing is $10$~m in both directions.
Velocity values are then linearly rescaled to $1500$--$4500$~m/s to match
the velocity range of the pretrained prior.
All three models use the same pretrained CurveFault-B EDM.
To apply this model to the larger fields, we divide each field into three
overlapping $70\times70$ patches and average the predictions in the
overlapping regions.
PSG retains its OpenFWI sampling and optimization settings on
Marmousi and Overthrust.
For Salt, preliminary reconstruction results from the compared methods
motivated staged evaluation of their solver hyperparameters and iteration
budgets; the selected PSG configuration uses two physical updates per
diffusion step.
The EDM remains fixed, and the solver parameters for each model are
held constant throughout its initialization and noise sensitivity tests.
Acquisition and patch adaptation are detailed in
Appendices~\ref{sec:appendix_solver} and~\ref{sec:appendix_baselines};
hyperparameter settings are given in Appendix~\ref{sec:exp_hyperparameters}.

\subsubsection{Reconstruction of Out-of-Distribution Velocity Models}
\label{sec:exp_ood_reconstruction}

\begin{figure}[!htbp]
  \centering
  \includegraphics[width=\linewidth]{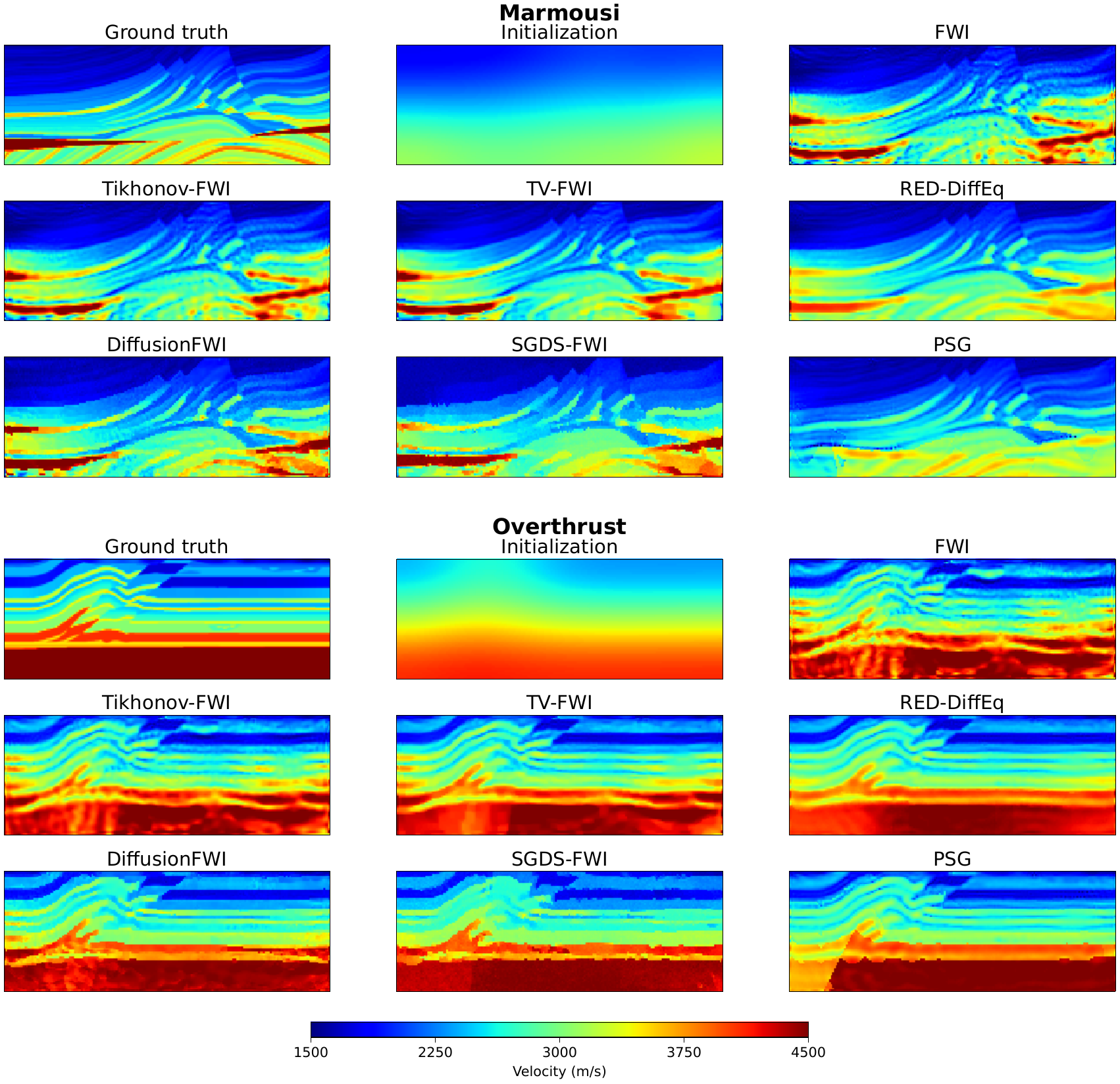}
  \caption{\textbf{Marmousi and Overthrust clean-data results.}
  Ground truth, smoothed initialization, and reconstructions from the
  compared methods at $\sigma_{\mathrm{init}}=20$.}
  \label{fig:ood_marmousi_overthrust}
\end{figure}

\begin{figure}[!htbp]
  \centering
  \includegraphics[width=\linewidth]{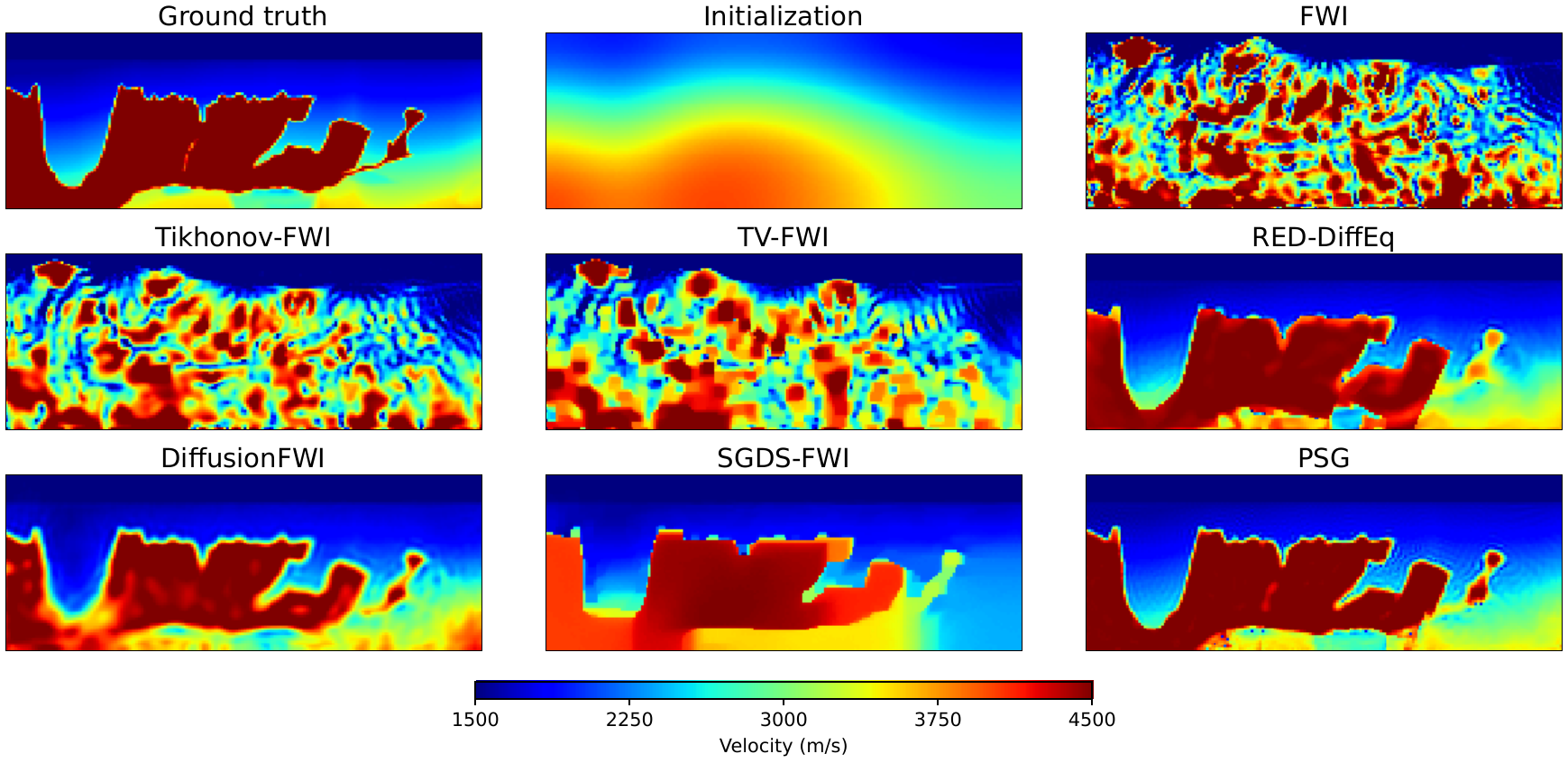}
  \caption{\textbf{BP2004 Salt clean-data results.}
  Ground truth, smoothed initialization, and reconstructions from the
  compared methods at $\sigma_{\mathrm{init}}=20$.}
  \label{fig:ood_salt}
\end{figure}

We first use clean observations and Gaussian-smoothed initial models
with $\sigma_{\mathrm{init}}=20$.
Fig.~\ref{fig:ood_marmousi_overthrust} and Fig.~\ref{fig:ood_salt}
show the ground truth, initialization, and reconstructed velocity fields.
The results of FWI, Tikhonov-FWI, TV-FWI, RED-DiffEq, DiffusionFWI, and
SGDS-FWI are retained alongside PSG under the same observations and
initial models.  In Fig.~\ref{fig:ood_marmousi_overthrust}, FWI,
Tikhonov-FWI, TV-FWI, and RED-DiffEq use $300$ optimization updates,
as in the OpenFWI experiment.

On Marmousi, PSG recovers the dominant curved layers and fault-related
discontinuities that are absent from the smoothed initialization.
The reconstructed field follows the main structural trends, although
thin deep layers and local velocity contrasts remain imperfectly resolved.
On Overthrust, PSG recovers the displaced layer geometry and the broad
high-velocity region at depth, while local interface positions and
velocity amplitudes still deviate from the ground truth.
The respective RMSE values are $331.6$ and $216.2$~m/s, with SSIM
values of $0.745$ and $0.825$.
These reconstructions show that the OpenFWI PSG configuration transfers
to both targets and recovers substantial structural detail beyond the
initial smooth background.

The Salt reconstruction provides a further test of transfer to a
geometry distinct from the layered training examples.
The initial model contains a broad velocity anomaly with little of the
true salt outline.  PSG recovers the major salt body together with
irregular upper boundaries and narrow geometric features
(Fig.~\ref{fig:ood_salt}).
Its MAE, RMSE, and SSIM are $43.9$~m/s, $123.1$~m/s, and $0.956$,
respectively, showing close agreement with the target under this
clean-data setting.  With the solver parameters selected on the Salt
case, the frozen CurveFault-B EDM thus supports recovery of a salt
geometry that differs substantially from its training structures.

\FloatBarrier
\subsubsection{Sensitivity to Initialization and Observation Noise}
\label{sec:exp_ood_sensitivity}

\begin{figure}[!htbp]
  \centering
  \includegraphics[width=\linewidth]{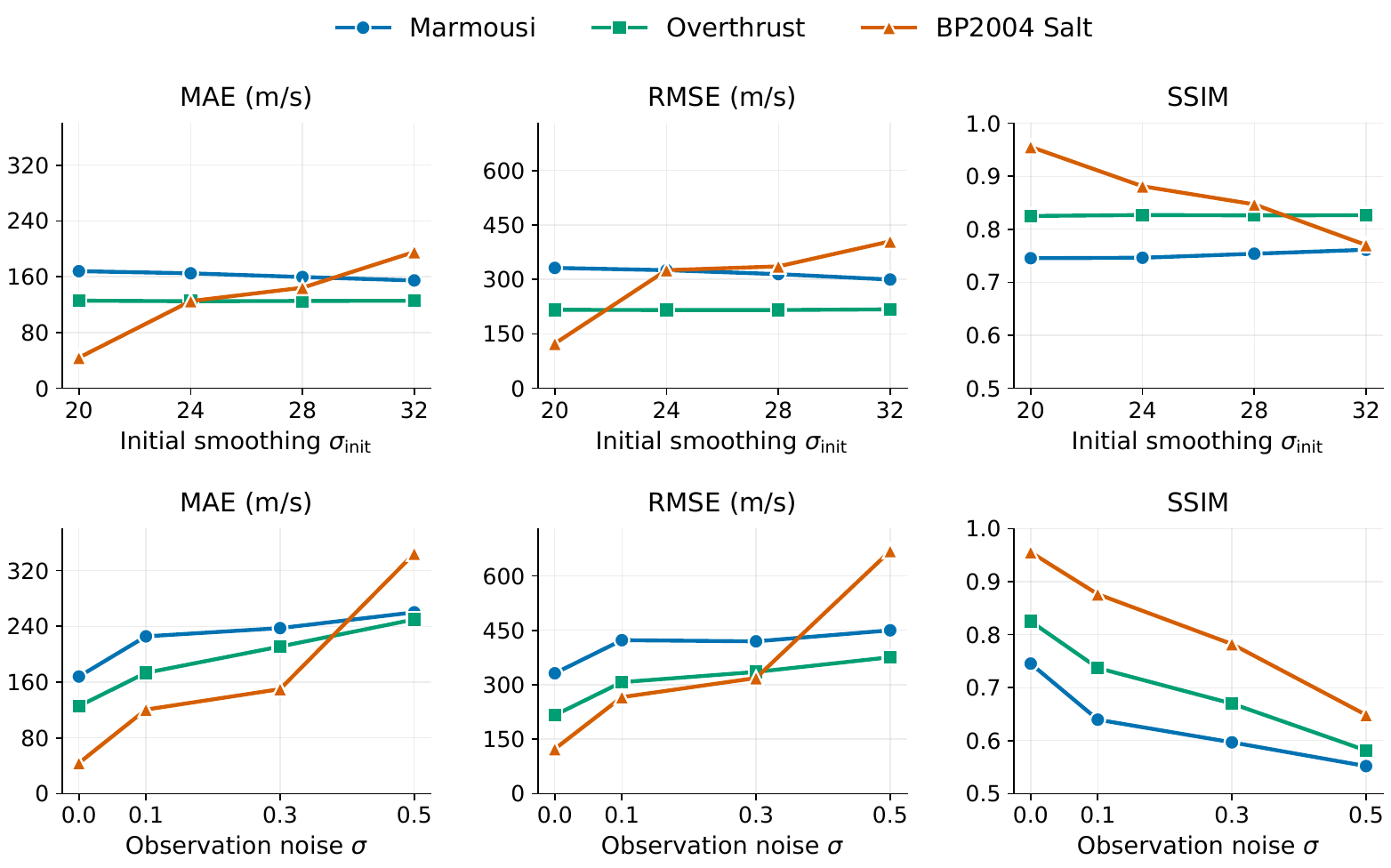}
  \caption{\textbf{PSG sensitivity on out-of-distribution models.}
  MAE, RMSE, and SSIM versus initialization smoothing on clean data
  (top) and Gaussian observation noise at $\sigma_{\mathrm{init}}=20$
  (bottom). Each point represents one reconstruction; the zero-noise
  values reuse the clean reference results.}
  \label{fig:ood_psg_sensitivity}
\end{figure}

\begin{figure}[p]
  \centering
  \includegraphics[width=\linewidth]{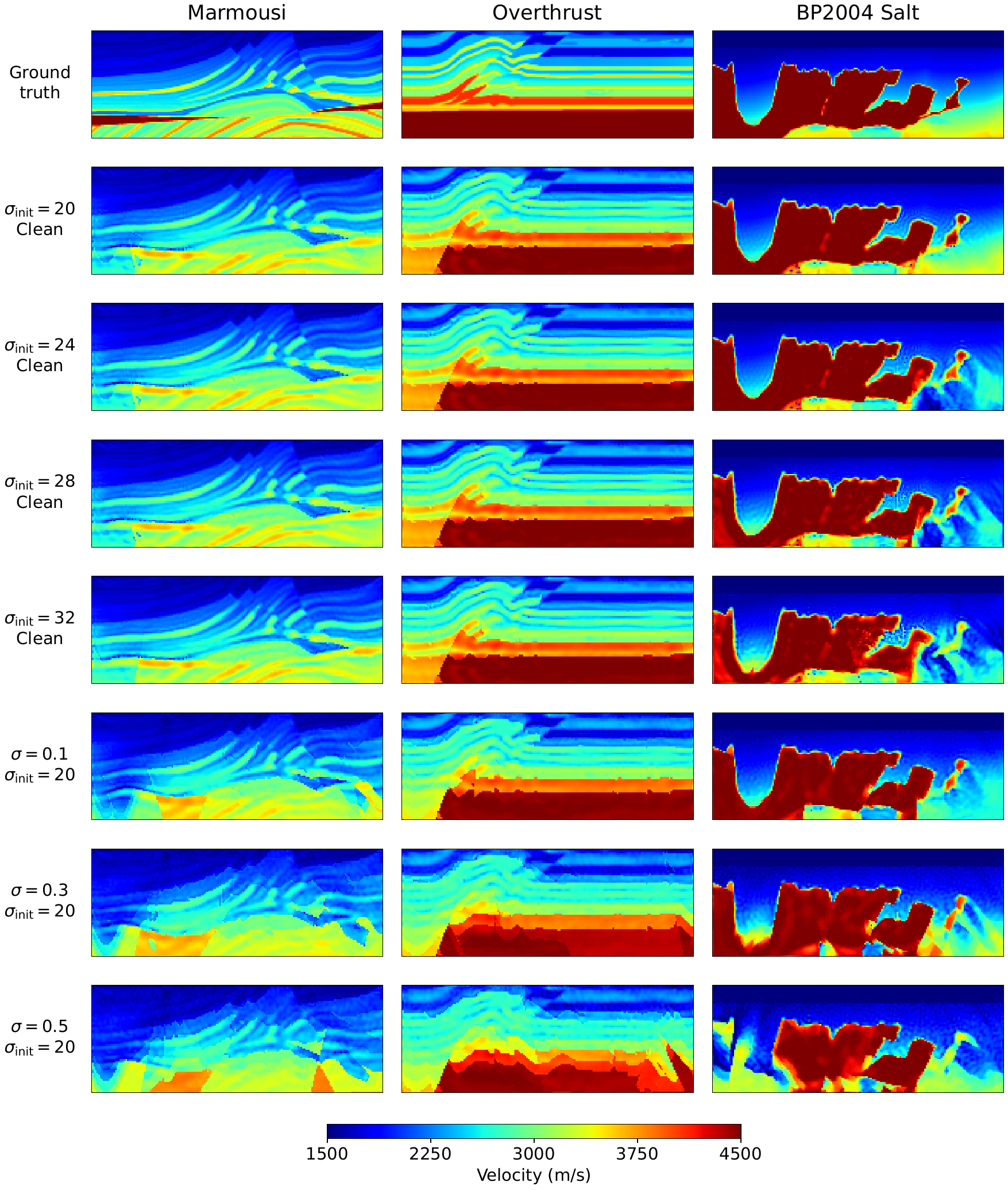}
  \caption{\textbf{PSG reconstructions under OOD initialization and noise variations.}
  Columns show Marmousi, Overthrust, and BP2004 Salt. Below the ground
  truth, rows show four initialization-smoothing levels on clean data
  and three Gaussian-noise levels at $\sigma_{\mathrm{init}}=20$.
  All panels share the same velocity scale.}
  \label{fig:ood_psg_reconstructions}
\end{figure}

We next examine how the OOD reconstructions depend on the initial model
and the quality of the observations.
For initialization sensitivity, we use clean data and increase the
Gaussian-smoothing scale through
$\sigma_{\mathrm{init}}\in\{20,24,28,32\}$.
For noise sensitivity, we fix $\sigma_{\mathrm{init}}=20$ and add
Gaussian observation noise with $\sigma\in\{0.1,0.3,0.5\}$, retaining
the clean case as the zero-noise reference.
Fig.~\ref{fig:ood_psg_sensitivity} plots the PSG metrics for all three
models under the fixed settings described above.
Fig.~\ref{fig:ood_psg_reconstructions} shows the corresponding physical
reconstructions alongside the ground truth for each model, covering all
seven initialization and noise conditions.

The initialization sweep shows limited sensitivity on Marmousi and
Overthrust.  On Marmousi, RMSE decreases from $331.6$ to $299.6$~m/s
across the tested smoothing scales, while SSIM changes from $0.745$ to
$0.761$.  On Overthrust, RMSE remains between $215.5$ and $217.4$~m/s
and SSIM between $0.825$ and $0.827$.
Salt is more sensitive to the loss of information in the initial model:
RMSE increases from $123.1$ to $404.7$~m/s and SSIM decreases from
$0.956$ to $0.770$ as the smoothing scale grows from $20$ to $32$.
The physical reconstructions in Fig.~\ref{fig:ood_psg_reconstructions}
show that the salt body remains recognizable, while
its boundary positions and surrounding velocities become less accurate.

Increasing observation noise generally raises the velocity errors and
reduces structural agreement on all three targets.
On Salt, PSG retains an SSIM of $0.876$ at $\sigma=0.1$ and $0.782$
at $\sigma=0.3$, but the strongest noise introduces substantial boundary
and background distortion.  Marmousi and Overthrust likewise lose
reconstruction accuracy as the noise increases, reaching SSIM values
of $0.552$ and $0.582$, respectively, at $\sigma=0.5$.

Taken together, these experiments demonstrate PSG's ability to reconstruct
velocity fields with complex, irregular geometries beyond the OpenFWI
training distribution.
PSG combines accurate recovery of the principal structures with
robustness to moderate observation noise and limited sensitivity to
initialization.
These findings support its direct generalization to new inversion
scenarios through reuse of a frozen diffusion prior, highlighting the
capacity of physical-state-guided sampling to recover unfamiliar
velocity structures without retraining the diffusion model.

\FloatBarrier

%% file: tables/openfwi_clean_metrics.tex
% Generated by reviews/openfwi_denoised_2026_09_11/build_review.py; PSG uses denoised output.
\begin{table}[!htbp]
  \centering
  \footnotesize
  \setlength{\tabcolsep}{2.5pt}
  \renewcommand{\arraystretch}{1.0}
  \caption{OpenFWI results with clean seismic data, averaged over 100 test samples
  per family. MAE and RMSE are in m/s. Bold indicates the best mean
  for each metric and family.}
  \label{tab:openfwi_clean_metrics}
  \begin{tabular*}{\linewidth}{@{\extracolsep{\fill}}l*{12}{r}@{}}
    \toprule
    Method & \multicolumn{3}{c}{\textbf{FV-B}} & \multicolumn{3}{c}{\textbf{FF-B}} & \multicolumn{3}{c}{\textbf{CV-B}} & \multicolumn{3}{c}{\textbf{CF-B}} \\
    \cmidrule(lr){2-4}\cmidrule(lr){5-7}\cmidrule(lr){8-10}\cmidrule(l){11-13}
    & MAE & RMSE & SSIM & MAE & RMSE & SSIM & MAE & RMSE & SSIM & MAE & RMSE & SSIM \\
    \midrule
    FWI & 214.9 & 338.7 & 0.681 & 175.0 & 269.0 & 0.696 & 350.6 & 531.5 & 0.522 & 263.7 & 400.8 & 0.603 \\
    Tikhonov-FWI & 207.4 & 328.1 & 0.702 & 168.3 & 259.7 & 0.730 & 339.6 & 515.9 & 0.543 & 252.1 & 384.1 & 0.633 \\
    TV-FWI & 198.3 & 320.7 & 0.741 & 156.7 & 247.7 & 0.783 & 328.3 & 507.0 & 0.580 & 241.3 & 372.3 & 0.669 \\
    DPS & 671.2 & 895.7 & 0.556 & 469.8 & 649.2 & 0.536 & 718.9 & 976.8 & 0.387 & 495.5 & 698.8 & 0.421 \\
    RED-DiffEq & 167.0 & 280.4 & 0.797 & 135.3 & 216.7 & 0.841 & 263.7 & 418.9 & 0.673 & 174.1 & 276.2 & 0.774 \\
    DiffusionFWI & 144.3 & 249.3 & 0.830 & 141.2 & 243.1 & 0.798 & 300.0 & 490.5 & 0.634 & 200.1 & 332.0 & 0.715 \\
    SGDS-FWI & 173.6 & 289.4 & 0.803 & 152.1 & 254.1 & 0.796 & 311.6 & 505.1 & 0.628 & 210.1 & 346.6 & 0.720 \\
    PSG & \textbf{110.2} & \textbf{209.6} & \textbf{0.880} & \textbf{84.9} & \textbf{166.4} & \textbf{0.904} & \textbf{188.1} & \textbf{357.3} & \textbf{0.789} & \textbf{112.0} & \textbf{219.1} & \textbf{0.860} \\
    \bottomrule
  \end{tabular*}
\end{table}

%% file: tables/openfwi_noise_metrics.tex
% Generated by reviews/openfwi_denoised_2026_09_11/build_review.py; PSG uses denoised output.
\begin{table}[p]
  \centering
  \footnotesize
  \setlength{\tabcolsep}{3.5pt}
  \renewcommand{\arraystretch}{1.0}
  \caption{OpenFWI results under Gaussian noise, averaged over 100 test samples
  per family. MAE and RMSE are in m/s. Bold indicates the best mean
  for each metric, family, and noise level.}
  \label{tab:openfwi_noise_metrics}
  \begin{tabular*}{\linewidth}{@{\extracolsep{\fill}}l*{9}{r}@{}}
    \toprule
    \multicolumn{10}{c}{\textbf{FV-B}} \\
    \midrule
    Method & \multicolumn{3}{c}{$\sigma=0.1$} & \multicolumn{3}{c}{$\sigma=0.3$} & \multicolumn{3}{c}{$\sigma=0.5$} \\
    \cmidrule(lr){2-4}\cmidrule(lr){5-7}\cmidrule(l){8-10}
    & MAE & RMSE & SSIM & MAE & RMSE & SSIM & MAE & RMSE & SSIM \\
    \midrule
    FWI & 245.3 & 373.3 & 0.604 & 288.1 & 422.7 & 0.488 & 323.5 & 466.0 & 0.415 \\
    Tikhonov-FWI & 230.6 & 353.2 & 0.655 & 259.5 & 383.7 & 0.569 & 278.7 & 403.2 & 0.506 \\
    TV-FWI & 214.3 & 338.9 & 0.731 & 232.4 & 357.8 & 0.704 & 241.5 & 366.0 & 0.676 \\
    DPS & 667.3 & 890.7 & 0.563 & 668.9 & 891.3 & 0.559 & 684.3 & 904.0 & 0.543 \\
    RED-DiffEq & 187.7 & 308.1 & 0.782 & 211.5 & 333.3 & 0.760 & 225.4 & \textbf{347.8} & 0.745 \\
    DiffusionFWI & 161.4 & 272.2 & 0.812 & 212.1 & 341.3 & 0.668 & 273.1 & 423.0 & 0.521 \\
    SGDS-FWI & 177.1 & 292.5 & 0.796 & 212.2 & 334.2 & 0.690 & 273.0 & 411.8 & 0.549 \\
    PSG & \textbf{128.9} & \textbf{240.7} & \textbf{0.857} & \textbf{179.8} & \textbf{316.5} & \textbf{0.812} & \textbf{206.5} & 352.5 & \textbf{0.778} \\
    \bottomrule
  \end{tabular*}
  \par\vspace{0.45em}
  \begin{tabular*}{\linewidth}{@{\extracolsep{\fill}}l*{9}{r}@{}}
    \toprule
    \multicolumn{10}{c}{\textbf{FF-B}} \\
    \midrule
    Method & \multicolumn{3}{c}{$\sigma=0.1$} & \multicolumn{3}{c}{$\sigma=0.3$} & \multicolumn{3}{c}{$\sigma=0.5$} \\
    \cmidrule(lr){2-4}\cmidrule(lr){5-7}\cmidrule(l){8-10}
    & MAE & RMSE & SSIM & MAE & RMSE & SSIM & MAE & RMSE & SSIM \\
    \midrule
    FWI & 217.8 & 323.6 & 0.595 & 267.2 & 386.8 & 0.478 & 306.8 & 439.0 & 0.409 \\
    Tikhonov-FWI & 201.0 & 299.8 & 0.666 & 232.1 & 335.3 & 0.579 & 255.5 & 362.8 & 0.517 \\
    TV-FWI & 183.1 & 279.9 & 0.752 & 202.6 & 300.2 & 0.712 & \textbf{213.1} & 310.9 & 0.681 \\
    DPS & 463.6 & 641.2 & 0.538 & 448.7 & 624.6 & 0.537 & 449.5 & 624.3 & 0.534 \\
    RED-DiffEq & 170.5 & 259.7 & 0.804 & 200.9 & \textbf{293.2} & 0.762 & 217.8 & \textbf{310.8} & 0.738 \\
    DiffusionFWI & 166.8 & 273.3 & 0.755 & 234.4 & 369.3 & 0.592 & 308.3 & 469.6 & 0.478 \\
    SGDS-FWI & 155.7 & 257.8 & 0.781 & 201.3 & 320.9 & 0.636 & 266.5 & 411.0 & 0.519 \\
    PSG & \textbf{127.8} & \textbf{227.5} & \textbf{0.867} & \textbf{194.0} & 321.7 & \textbf{0.800} & 228.9 & 366.5 & \textbf{0.764} \\
    \bottomrule
  \end{tabular*}
  \par\vspace{0.45em}
  \begin{tabular*}{\linewidth}{@{\extracolsep{\fill}}l*{9}{r}@{}}
    \toprule
    \multicolumn{10}{c}{\textbf{CV-B}} \\
    \midrule
    Method & \multicolumn{3}{c}{$\sigma=0.1$} & \multicolumn{3}{c}{$\sigma=0.3$} & \multicolumn{3}{c}{$\sigma=0.5$} \\
    \cmidrule(lr){2-4}\cmidrule(lr){5-7}\cmidrule(l){8-10}
    & MAE & RMSE & SSIM & MAE & RMSE & SSIM & MAE & RMSE & SSIM \\
    \midrule
    FWI & 372.9 & 553.5 & 0.475 & 400.3 & 581.0 & 0.414 & 425.8 & 608.4 & 0.369 \\
    Tikhonov-FWI & 355.4 & 529.2 & 0.513 & 371.5 & 540.8 & 0.468 & 384.2 & 549.8 & 0.433 \\
    TV-FWI & 338.9 & 516.2 & 0.570 & 348.0 & 520.8 & 0.552 & 354.8 & 523.0 & 0.533 \\
    DPS & 714.6 & 972.1 & 0.389 & 712.8 & 966.9 & 0.388 & 710.0 & 959.5 & 0.390 \\
    RED-DiffEq & 280.9 & 434.3 & 0.664 & 311.6 & \textbf{458.8} & 0.638 & 332.9 & \textbf{474.9} & 0.618 \\
    DiffusionFWI & 317.4 & 512.6 & 0.607 & 350.5 & 552.2 & 0.530 & 396.2 & 609.3 & 0.449 \\
    SGDS-FWI & 315.1 & 507.5 & 0.621 & 347.7 & 542.9 & 0.553 & 402.1 & 609.2 & 0.471 \\
    PSG & \textbf{204.2} & \textbf{388.7} & \textbf{0.781} & \textbf{269.0} & 480.8 & \textbf{0.718} & \textbf{325.5} & 557.6 & \textbf{0.670} \\
    \bottomrule
  \end{tabular*}
  \par\vspace{0.45em}
  \begin{tabular*}{\linewidth}{@{\extracolsep{\fill}}l*{9}{r}@{}}
    \toprule
    \multicolumn{10}{c}{\textbf{CF-B}} \\
    \midrule
    Method & \multicolumn{3}{c}{$\sigma=0.1$} & \multicolumn{3}{c}{$\sigma=0.3$} & \multicolumn{3}{c}{$\sigma=0.5$} \\
    \cmidrule(lr){2-4}\cmidrule(lr){5-7}\cmidrule(l){8-10}
    & MAE & RMSE & SSIM & MAE & RMSE & SSIM & MAE & RMSE & SSIM \\
    \midrule
    FWI & 303.9 & 450.8 & 0.529 & 340.3 & 493.6 & 0.449 & 369.1 & 529.9 & 0.396 \\
    Tikhonov-FWI & 285.8 & 424.2 & 0.576 & 311.4 & 451.2 & 0.515 & 326.8 & 467.1 & 0.473 \\
    TV-FWI & 268.0 & 404.8 & 0.632 & 284.5 & 419.4 & 0.594 & 290.8 & 423.1 & 0.570 \\
    DPS & 494.2 & 696.5 & 0.421 & 487.2 & 682.4 & 0.418 & 482.5 & 671.4 & 0.413 \\
    RED-DiffEq & 210.3 & 320.3 & 0.726 & 242.9 & \textbf{354.2} & 0.679 & \textbf{260.8} & \textbf{371.1} & 0.653 \\
    DiffusionFWI & 242.4 & 387.7 & 0.658 & 297.8 & 464.1 & 0.555 & 360.3 & 546.3 & 0.466 \\
    SGDS-FWI & 214.0 & 350.5 & 0.708 & 246.8 & 396.5 & 0.619 & 304.4 & 475.4 & 0.527 \\
    PSG & \textbf{160.2} & \textbf{294.4} & \textbf{0.806} & \textbf{238.6} & 400.2 & \textbf{0.720} & 293.7 & 470.2 & \textbf{0.660} \\
    \bottomrule
  \end{tabular*}
\end{table}

%% file: tables/openfwi_missing_metrics.tex
% Generated by reviews/openfwi_denoised_2026_09_11/build_review.py; PSG uses denoised output.
\begin{table}[p]
  \centering
  \footnotesize
  \setlength{\tabcolsep}{3.5pt}
  \renewcommand{\arraystretch}{1.0}
  \caption{OpenFWI results with missing traces, averaged over 100 test samples
  per family. MAE and RMSE are in m/s. Bold indicates the best mean
  for each metric, family, and missing-trace level.}
  \label{tab:openfwi_missing_metrics}
  \label{tab:missing_results}
  \begin{tabular*}{\linewidth}{@{\extracolsep{\fill}}l*{9}{r}@{}}
    \toprule
    \multicolumn{10}{c}{\textbf{FV-B}} \\
    \midrule
    Method & \multicolumn{3}{c}{10/70 missing} & \multicolumn{3}{c}{35/70 missing} & \multicolumn{3}{c}{60/70 missing} \\
    \cmidrule(lr){2-4}\cmidrule(lr){5-7}\cmidrule(l){8-10}
    & MAE & RMSE & SSIM & MAE & RMSE & SSIM & MAE & RMSE & SSIM \\
    \midrule
    FWI & 216.1 & 339.9 & 0.679 & 219.2 & 343.2 & 0.673 & 229.5 & 351.8 & 0.640 \\
    Tikhonov-FWI & 208.4 & 329.0 & 0.701 & 211.2 & 332.1 & 0.696 & 221.6 & 341.2 & 0.662 \\
    TV-FWI & 199.4 & 321.5 & 0.739 & 202.0 & 324.6 & 0.735 & 210.8 & 332.6 & 0.710 \\
    DPS & 648.0 & 867.7 & 0.575 & 635.2 & 849.4 & 0.562 & 618.8 & 834.9 & 0.565 \\
    RED-DiffEq & 168.0 & 281.4 & 0.795 & 168.8 & 281.1 & 0.794 & 175.0 & 288.0 & 0.778 \\
    DiffusionFWI & 145.2 & 251.2 & 0.829 & 146.7 & 253.6 & 0.828 & 151.5 & 258.5 & 0.819 \\
    SGDS-FWI & 176.0 & 292.4 & 0.801 & 177.6 & 294.5 & 0.801 & 176.0 & 292.1 & 0.799 \\
    PSG & \textbf{108.8} & \textbf{207.6} & \textbf{0.881} & \textbf{108.1} & \textbf{207.4} & \textbf{0.880} & \textbf{115.8} & \textbf{219.4} & \textbf{0.871} \\
    \bottomrule
  \end{tabular*}
  \par\vspace{0.45em}
  \begin{tabular*}{\linewidth}{@{\extracolsep{\fill}}l*{9}{r}@{}}
    \toprule
    \multicolumn{10}{c}{\textbf{FF-B}} \\
    \midrule
    Method & \multicolumn{3}{c}{10/70 missing} & \multicolumn{3}{c}{35/70 missing} & \multicolumn{3}{c}{60/70 missing} \\
    \cmidrule(lr){2-4}\cmidrule(lr){5-7}\cmidrule(l){8-10}
    & MAE & RMSE & SSIM & MAE & RMSE & SSIM & MAE & RMSE & SSIM \\
    \midrule
    FWI & 177.2 & 271.6 & 0.693 & 181.5 & 276.1 & 0.683 & 216.0 & 315.9 & 0.594 \\
    Tikhonov-FWI & 169.9 & 261.6 & 0.727 & 173.4 & 265.3 & 0.718 & 206.6 & 303.5 & 0.627 \\
    TV-FWI & 158.1 & 249.4 & 0.781 & 161.0 & 252.2 & 0.773 & 191.6 & 287.4 & 0.693 \\
    DPS & 460.4 & 636.9 & 0.537 & 444.9 & 622.4 & 0.544 & 458.2 & 632.6 & 0.518 \\
    RED-DiffEq & 136.5 & 217.7 & 0.839 & 140.4 & 222.1 & 0.832 & 161.0 & 244.7 & 0.787 \\
    DiffusionFWI & 141.2 & 242.3 & 0.798 & 142.7 & 244.0 & 0.795 & 160.1 & 262.5 & 0.758 \\
    SGDS-FWI & 152.8 & 255.1 & 0.795 & 156.5 & 259.6 & 0.789 & 175.7 & 282.4 & 0.747 \\
    PSG & \textbf{86.1} & \textbf{168.0} & \textbf{0.902} & \textbf{89.4} & \textbf{171.9} & \textbf{0.899} & \textbf{111.7} & \textbf{200.7} & \textbf{0.871} \\
    \bottomrule
  \end{tabular*}
  \par\vspace{0.45em}
  \begin{tabular*}{\linewidth}{@{\extracolsep{\fill}}l*{9}{r}@{}}
    \toprule
    \multicolumn{10}{c}{\textbf{CV-B}} \\
    \midrule
    Method & \multicolumn{3}{c}{10/70 missing} & \multicolumn{3}{c}{35/70 missing} & \multicolumn{3}{c}{60/70 missing} \\
    \cmidrule(lr){2-4}\cmidrule(lr){5-7}\cmidrule(l){8-10}
    & MAE & RMSE & SSIM & MAE & RMSE & SSIM & MAE & RMSE & SSIM \\
    \midrule
    FWI & 349.6 & 530.1 & 0.523 & 352.4 & 532.4 & 0.516 & 372.4 & 543.9 & 0.473 \\
    Tikhonov-FWI & 338.8 & 514.7 & 0.544 & 341.5 & 516.6 & 0.537 & 362.1 & 529.0 & 0.491 \\
    TV-FWI & 327.6 & 506.4 & 0.580 & 330.7 & 508.9 & 0.573 & 351.3 & 519.9 & 0.527 \\
    DPS & 701.5 & 957.7 & 0.394 & 691.5 & 943.5 & 0.401 & 681.2 & 911.0 & 0.392 \\
    RED-DiffEq & 263.6 & 418.0 & 0.673 & 267.1 & 422.0 & 0.667 & 286.5 & 437.2 & 0.631 \\
    DiffusionFWI & 300.8 & 491.6 & 0.632 & 299.6 & 489.2 & 0.631 & 312.8 & 497.9 & 0.608 \\
    SGDS-FWI & 312.2 & 506.7 & 0.628 & 312.1 & 507.0 & 0.626 & 336.8 & 531.1 & 0.595 \\
    PSG & \textbf{187.5} & \textbf{357.2} & \textbf{0.790} & \textbf{187.8} & \textbf{356.5} & \textbf{0.789} & \textbf{211.9} & \textbf{379.7} & \textbf{0.757} \\
    \bottomrule
  \end{tabular*}
  \par\vspace{0.45em}
  \begin{tabular*}{\linewidth}{@{\extracolsep{\fill}}l*{9}{r}@{}}
    \toprule
    \multicolumn{10}{c}{\textbf{CF-B}} \\
    \midrule
    Method & \multicolumn{3}{c}{10/70 missing} & \multicolumn{3}{c}{35/70 missing} & \multicolumn{3}{c}{60/70 missing} \\
    \cmidrule(lr){2-4}\cmidrule(lr){5-7}\cmidrule(l){8-10}
    & MAE & RMSE & SSIM & MAE & RMSE & SSIM & MAE & RMSE & SSIM \\
    \midrule
    FWI & 262.4 & 398.6 & 0.603 & 266.9 & 403.6 & 0.593 & 319.8 & 460.6 & 0.484 \\
    Tikhonov-FWI & 251.2 & 382.6 & 0.632 & 255.6 & 387.5 & 0.621 & 308.0 & 444.0 & 0.508 \\
    TV-FWI & 239.0 & 368.9 & 0.672 & 242.3 & 372.6 & 0.660 & 294.1 & 427.7 & 0.547 \\
    DPS & 494.4 & 695.2 & 0.417 & 479.7 & 671.7 & 0.420 & 484.6 & 675.0 & 0.407 \\
    RED-DiffEq & 174.0 & 275.8 & 0.773 & 179.3 & 282.6 & 0.764 & 221.0 & 328.1 & 0.673 \\
    DiffusionFWI & 199.9 & 331.8 & 0.716 & 200.0 & 331.4 & 0.713 & 241.8 & 379.8 & 0.639 \\
    SGDS-FWI & 208.7 & 345.6 & 0.720 & 213.5 & 349.6 & 0.709 & 251.7 & 392.7 & 0.645 \\
    PSG & \textbf{112.9} & \textbf{220.5} & \textbf{0.859} & \textbf{113.3} & \textbf{220.1} & \textbf{0.858} & \textbf{143.9} & \textbf{262.0} & \textbf{0.820} \\
    \bottomrule
  \end{tabular*}
\end{table}

%% file: content/conclusion.tex
\section{Conclusion}
\label{sec:conclusion}

We presented Physical-State-Guided Diffusion Sampling (PSG), a framework
that integrates a pretrained diffusion prior into full-waveform inversion
through a persistent physical velocity state.  By coupling physical
updates driven by waveform consistency with diffusion sampling, PSG
accommodates flexible physical initialization and separates the nonlinear
forward operator from the denoiser in gradient computation.

The terminal denoised estimates on four OpenFWI families recover sharp
geological structures, with the best mean reconstruction metrics under clean and
missing-trace acquisitions and strong structural recovery under measurement
noise.  Repeated stochastic runs retain the dominant geological structures,
and the illustrated ensemble means remain close to the ground-truth fields.
Variability is concentrated primarily near faults and geological interfaces
and is positively associated with local inversion error across most test
models.  Experiments on Marmousi, Overthrust, and BP2004
Salt further show that a single OpenFWI-trained EDM prior can support
inversion of larger models with more complex geological structures
without retraining.

The physical-state-guided formulation offers
a general approach to combining learned priors with nonlinear PDE
constraints.  Future work will focus on more realistic geological priors
and acquisition settings, and on extending the framework to other
PDE-governed inverse problems.

%% file: content/appendix.tex
\input{content/appendix_diffusion}

\input{content/appendix_implementation}

%% file: content/appendix_diffusion.tex
\section{Diffusion Model Preliminaries}
\label{sec:appendix_diffusion_preliminaries}

Recall the noisy family defined in Section~\ref{sec:diffusion_prelim},
$x_t=a_tx_0+b_t\epsilon$ from Eq.~\eqref{eq:vp_corruption}, where $x_0\sim p_0$,
$\epsilon\sim\N(0,I)$, $a_t=\sqrt{\bar\alpha_t}$, and
$b_t=\sqrt{1-\bar\alpha_t}$.  Let $\varphi_b$ denote the density of
$\N(0,b^2I)$.  The density of $x_t$ is then given by the convolution
\begin{equation*}
  p_t=\bigl[a_t^{-d}p_0(\,\cdot\,/a_t)\bigr]*\varphi_{b_t}.
\end{equation*}
Following the Fourier-domain derivation of Karras et al.~\cite[Appendix B.5.1]{karras2022elucidating},
we obtain the transformed density and its time derivative:
\begin{equation*}
  \begin{aligned}
    \widehat p_t(\xi)
      &=\widehat p_0(a_t\xi)
        \exp\!\left(-\frac{b_t^2}{2}\Norm{\xi}^2\right),\\
    \partial_t\widehat p_t(\xi)
      &=\frac{\dot a_t}{a_t}\,\xi\!\cdot\!\nabla_\xi\widehat p_t(\xi)
        -\left(b_t\dot b_t-\frac{\dot a_t}{a_t}b_t^2\right)
          \Norm{\xi}^2\widehat p_t(\xi).
  \end{aligned}
\end{equation*}
Transforming back to $x$ expresses the density evolution in terms of the
corruption coefficients:
\begin{equation}
  \label{eq:appendix_density_evolution}
  \begin{aligned}
    \partial_t p_t(x)
      &=-\frac{\dot a_t}{a_t}\nabla_x\!\cdot\!\bigl(xp_t(x)\bigr)
        +\left(b_t\dot b_t-\frac{\dot a_t}{a_t}b_t^2\right)\Delta_xp_t(x)\\
      &=-\frac{\dot{\bar\alpha}_t}{2\bar\alpha_t}
        \left[\nabla_x\!\cdot\!\bigl(xp_t(x)\bigr)+\Delta_xp_t(x)\right].
  \end{aligned}
\end{equation}
The second equality follows from $a_t^2+b_t^2=1$, which gives
$b_t\dot b_t-(\dot a_t/a_t)b_t^2=-\dot a_t/a_t$.
The divergence term accounts for the changing signal scale, so the VP
density obeys a drift--diffusion equation in the original $x$ coordinates.

To construct a process with these marginals, consider the reverse-time SDE
$\mathrm dx_t=f(x_t,t)\,\mathrm dt+g(t)\,\mathrm d\bar w_t$,
with $t:T\to0$.  Here $\mathrm dt<0$, and the Brownian increment has
covariance $|\mathrm dt|I$.  Its density $\rho_t$ therefore satisfies the
Fokker--Planck equation in the decreasing $t$ coordinate~\cite{song2020score}
\begin{equation*}
  \partial_t\rho_t(x)
    =-\nabla_x\!\cdot\!\bigl(f(x,t)\rho_t(x)\bigr)
      -\frac{g(t)^2}{2}\Delta_x\rho_t(x).
\end{equation*}
To recover Eq.~\eqref{eq:appendix_density_evolution}, choose the drift
\begin{equation*}
  f(x,t)=\frac{\dot a_t}{a_t}x
    +\left(\frac{\dot a_t}{a_t}-\frac{g(t)^2}{2}\right)
      \nabla_x\log p_t(x).
\end{equation*}
Substituting $\rho_t=p_t$ and using
$\nabla_x\!\cdot\!\bigl(p_t\nabla_x\log p_t\bigr)=\Delta_xp_t$
recovers the prescribed density evolution.  Thus, for admissible $g(t)\geq0$
and terminal density $p_T$, the reverse process has marginals $p_t$.

Setting $g(t)=0$ gives the VP form of the probability-flow ODE
in Eq.~\eqref{eq:probability_flow_ode}
\begin{equation*}
  \frac{\mathrm dx_t}{\mathrm dt}
    =\frac{\dot a_t}{a_t}
      \left[x_t+\nabla_{x_t}\log p_t(x_t)\right].
\end{equation*}
Integrating this ODE from $T$ to $0$ transports $p_T$ back to $p_0$.
Under suitable time and state parameterizations, this flow underlies
deterministic samplers such as DDIM~\cite{song2020score,song2020denoising,karras2022elucidating}.

The variance-exploding (VE) parameterization used by EDM takes
$z_\sigma=x_0+\sigma\epsilon$, with density
$q_\sigma=p_0*\N(0,\sigma^2I)$~\cite{karras2022elucidating}.
The signal scale remains fixed while the noise variance increases.
At a fixed VP level $\bar\alpha_t$, its relation to the VE family is
\begin{equation}
  \label{eq:appendix_vp_ve_correspondence}
  \sigma^2=\frac{1-\bar\alpha_t}{\bar\alpha_t},
  \qquad z_\sigma=\frac{x_t}{a_t},
  \qquad \E[x_0\given x_t]=\E[x_0\given z_\sigma].
\end{equation}
The conditional means agree because the scaling is invertible at fixed $t$.
An EDM denoiser can therefore be used for VP sampling by rescaling its input
and supplying the matched noise level.

The score needed for sampling can be learned through denoising.  In VE
coordinates with $\sigma>0$, differentiating the Gaussian convolution gives Tweedie's
formula~\cite{efron2011tweedie,karras2022elucidating}:
\begin{equation}
  \label{eq:appendix_ve_tweedie}
  \nabla_z\log q_\sigma(z)
    =\frac{\int (x_0-z)\,p_0(x_0)\varphi_\sigma(z-x_0)\,\mathrm dx_0}
      {\sigma^2q_\sigma(z)}.
\end{equation}
At a fixed noise level $\sigma$, consider the denoising objective in
continuous form~\cite[Appendix B.3]{karras2022elucidating}:
\begin{equation*}
  \begin{aligned}
    \mathcal L[D;\sigma]
      &=\E_{x_0,\epsilon}\!\left[\Norm{D(x_0+\sigma\epsilon,\sigma)-x_0}^2\right]\\
      &=\int q_\sigma(z)\,
        \E\!\left[\Norm{D(z,\sigma)-x_0}^2\given z_\sigma=z\right]\mathrm dz\\
      &=\int\!\int p_0(x_0)\varphi_\sigma(z-x_0)
        \Norm{D(z,\sigma)-x_0}^2\,\mathrm dx_0\,\mathrm dz.
  \end{aligned}
\end{equation*}
Varying $D$ independently at each $z$ gives the stationarity condition
\begin{equation*}
  \begin{aligned}
    0
      &=\left.\frac{\delta\mathcal L[D;\sigma]}{\delta D(z,\sigma)}
        \right|_{D=D^\star}\\
      &=2\int p_0(x_0)\varphi_\sigma(z-x_0)
        \bigl(D^\star(z,\sigma)-x_0\bigr)\,\mathrm dx_0.
  \end{aligned}
\end{equation*}
Since $q_\sigma(z)>0$, the pointwise quadratic is strictly convex, and its
unique minimizer is
\begin{equation}
  \label{eq:appendix_optimal_denoiser}
  D^\star(z,\sigma)
    =\frac{\int x_0p_0(x_0)\varphi_\sigma(z-x_0)\,\mathrm dx_0}
      {\int p_0(x_0)\varphi_\sigma(z-x_0)\,\mathrm dx_0}
    =\E[x_0\given z_\sigma=z].
\end{equation}
The optimal denoiser in Eq.~\eqref{eq:appendix_optimal_denoiser}
thus supplies the unknown conditional mean needed in
Eq.~\eqref{eq:appendix_ve_tweedie}, yielding the score
$(D^\star(z,\sigma)-z)/\sigma^2$.  Its VP counterpart in
Eq.~\eqref{eq:tweedie_score} follows from
Eq.~\eqref{eq:appendix_vp_ve_correspondence}.
In practice, a neural denoiser minimizes
$\E_{\sigma\sim\nu}[\lambda(\sigma)\mathcal L[D_\theta;\sigma]]$,
where $\nu$ samples positive noise levels and $\lambda(\sigma)>0$.

These relations explain the basic mechanism of diffusion sampling: the
noisy family prescribes the density evolution, and matching its
Fokker--Planck equation yields reverse SDEs and an ODE driven by the score.
Finite samples from training datasets do not directly provide the density
$p_0$, instead they allow approximation of integrals with respect to $p_0$
by empirical averages.  Using a neural approximation of the score, diffusion
models numerically solve the reverse SDE or probability-flow ODE on a
discrete time grid, transforming initial Gaussian noise into approximate
samples from the data distribution.

%% file: content/appendix_implementation.tex
\section{Implementation Details}
\label{sec:appendix_implementation}

This appendix specifies prior pretraining, the acoustic solver, method
implementation protocols, hyperparameters, and computational cost.

\subsection{Diffusion Model Pretraining}
\label{sec:appendix_pretraining}

We independently train an unconditional EDM model for each OpenFWI
family~\cite{deng2022openfwi,karras2022elucidating}, using $24{,}000$
velocity fields for FlatVel-B and CurveVel-B and $48{,}000$ for
FlatFault-B and CurveFault-B.  Each family has a separate $6000$-field
monitoring split, which includes the inversion test files but is excluded
from gradient training.  Velocities are normalized by $x=(v-3000)/1500$,
with $v$ in m/s.  The diffusion model operates in these normalized velocity
coordinates during both training and inference.  Before each wave-equation
solve, the velocity supplied to the forward operator is converted back to
physical units by $v=1500x+3000$.

\paragraph{Network architecture.}
The denoiser is a Dhariwal U-Net~\cite{dhariwal2021diffusion} with one input and one output channel,
channel widths $(64,128,192,256)$, two residual blocks per level, and
attention at resolutions $36$, $18$, and $9$.  It has approximately
$25.4$ million parameters.  Native $70\times70$ fields are padded by
boundary replication to $72\times72$ and cropped back after denoising.

\begin{figure}[!htbp]
  \centering
  \includegraphics[width=0.80\linewidth]{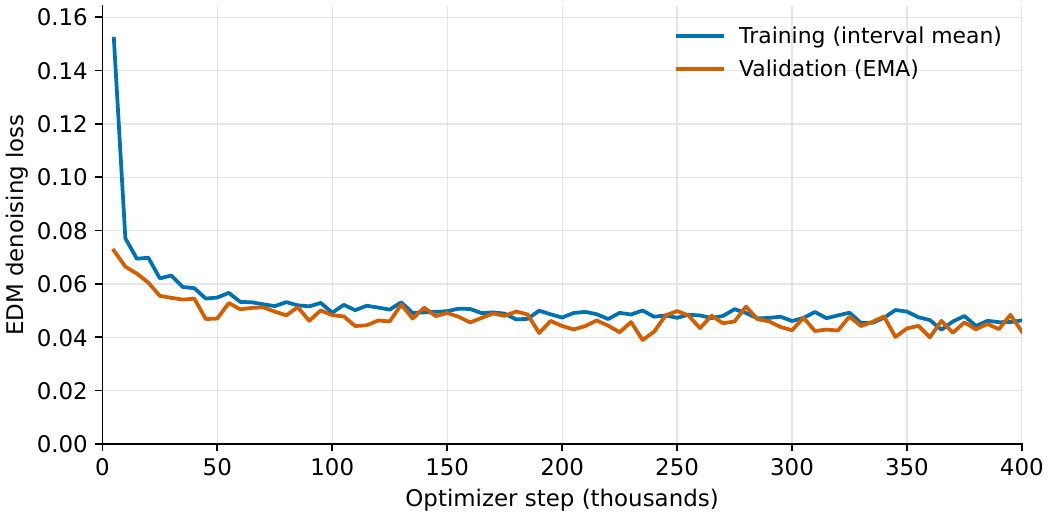}
  \caption{\textbf{EDM pretraining loss on CurveFault-B.}
  Training and validation weighted denoising losses versus optimizer step.
  The other three pretrained models exhibit qualitatively similar
  training and validation curves.}
  \label{fig:edm_pretraining_loss}
\end{figure}

\paragraph{Training settings.}
The noise-level distribution, loss weighting, and preconditioning follow
the official EDM implementation~\cite{karras2022elucidating}.\footnote{\url{https://github.com/NVlabs/edm}}
We train with Adam~\cite{kingma2015adam} at learning rate $10^{-4}$ and batch size $32$ for
$400{,}000$ updates, using a $500$-step warm-up and FP16 mixed precision.
The final EMA weights are used without selection by monitoring loss.
Fig.~\ref{fig:edm_pretraining_loss} shows the recorded training and
validation losses for CurveFault-B.  After roughly $3\times10^5$
optimizer steps, the loss oscillates around a slowly decreasing plateau.
Such fluctuations are typical of the stochastic EDM denoising objective,
in which noise levels and Gaussian perturbations are resampled for each
minibatch.  The empirical loss therefore retains sampling variability
even near convergence.

\paragraph{VP schedule for algorithm calls.}
Diffusion inverse solver baselines and PSG use VP noise coordinates
within the DDPM framework~\cite{ho2020denoising}.
Algorithm~\ref{alg:vp_schedule} specifies the discrete linear DDPM schedule
used by RED-DiffEq and DiffusionFWI.
The EDM denoiser is called with input $x_t/\sqrt{\bar\alpha_t}$ and noise
level $\sigma_t$, as in Eq.~\eqref{eq:appendix_vp_ve_correspondence}.
Method-specific schedules and discretizations are given in
Appendix~\ref{sec:exp_hyperparameters}.

\begin{algorithm}[H]
  \caption{Linear $\beta$ schedule under the DDPM framework}
  \label{alg:vp_schedule}
  \small
  \begin{algorithmic}[1]
    \REQUIRE Number of levels $T=1000$; initial variance $\beta_1=10^{-4}$;
      final variance $\beta_T=2\times10^{-2}$.
    \STATE $\bar\alpha_0\leftarrow1$
    \FOR{$t=1$ \textbf{to} $T$}
      \STATE $\beta_t\leftarrow\beta_1+
        \dfrac{t-1}{T-1}(\beta_T-\beta_1)$
      \STATE $\alpha_t\leftarrow1-\beta_t$
      \STATE $\bar\alpha_t\leftarrow\alpha_t\bar\alpha_{t-1}$
      \STATE $\sigma_t\leftarrow\sqrt{(1-\bar\alpha_t)/\bar\alpha_t}$
    \ENDFOR
    \RETURN $\{\beta_t,\alpha_t,\bar\alpha_t,\sigma_t\}_{t=1}^{T}$
  \end{algorithmic}
\end{algorithm}

DPS and PSG use the continuous VP schedule of the sampling implementation
~\cite{song2020score,zheng2025inversebench}, with rate and marginal coefficients
\begin{equation*}
  \begin{aligned}
    \beta(t)&=\beta_{\min}+t(\beta_{\max}-\beta_{\min})
      =0.1+19.9t, \qquad 0\leq t\leq1,\\
    \bar\alpha_t&=\exp\!\left(-\int_0^t\beta(s)\,\mathrm ds\right)
      =\exp(-0.1t-9.95t^2),\\
    \sigma_t&=\sqrt{\bar\alpha_t^{-1}-1},
  \end{aligned}
\end{equation*}
where $\beta_{\min}=0.1$, $\beta_{\max}=20$, and $\sigma_t$ is the
relative noise level passed to the EDM denoiser.
The $1000$ score-evaluation times are uniformly spaced from $1$ to $10^{-3}$;
the final update ends at the time corresponding to $\sigma_t=0.01$.
The continuous rate $\beta(t)$ determines the Euler--Maruyama coefficients
through $\beta(t_n)(t_n-t_{n-1})$.
For the ancestral representation in Eq.~\eqref{eq:diffusion_update}, the
discrete variance is instead
$\beta_{t_n}=1-\bar\alpha_{t_n}/\bar\alpha_{t_{n-1}}$.

\FloatBarrier

\subsection{Acoustic Wave-Equation Forward Solver}
\label{sec:appendix_solver}
\label{sec:exp_protocol}

The experimental forward solver integrates the two-dimensional,
constant-density acoustic wave equation using fourth-order spatial finite
differences and second-order time stepping.  A Ricker wavelet excites the
sources.  We retain the benchmark solver's numerical boundary treatment,
using a sponge layer with quadratic damping to reduce artificial reflections.
Sources are approximately uniformly
spaced horizontally, with receivers at every horizontal grid point; both
are placed at a fixed shallow depth.  OpenFWI and the OOD targets share
the numerical scheme and acquisition settings, with the horizontal extent
and receiver count increased for the wider models.  Velocity gradients
are computed by the discrete adjoint, which propagates waveform-residual
derivatives backward through the time-stepping scheme and combines them
with the forward wavefields~\cite{tarantola1984inversion,virieux2009overview}.
Table~\ref{tab:acoustic_geometry} summarizes the configuration.

\begin{table}[!htbp]
\centering\small
\caption{Forward-solver configuration. OOD targets comprise Marmousi,
Overthrust, and Salt. All models share the numerical scheme, source
wavelet, and boundary treatment; the horizontal grid size and receiver
count increase for the wider targets. Grid footprints are depth by width.}
\label{tab:acoustic_geometry}
\begin{tabular}{lcccc}
\toprule
Parameter & Symbol & OpenFWI & OOD targets & Unit \\
\midrule
Horizontal grid points & $m_x$ & $70$ & $190$ & --- \\
Vertical grid points & $m_z$ & $70$ & $70$ & --- \\
Grid spacing & $\Delta x,\Delta z$ & $10.0$ & $10.0$ & m \\
Time samples & $N_t$ & $1000$ & $1000$ & --- \\
Time step size & $\Delta t$ & $0.001$ & $0.001$ & s \\
Recording time & $T$ & $1.0$ & $1.0$ & s \\
Source frequency (Ricker) & $f$ & $15.0$ & $15.0$ & Hz \\
Number of sources & $N_s$ & $5$ & $5$ & --- \\
Number of receivers & $N_r$ & $70$ & $190$ & --- \\
Source depth & $z_s$ & $10$ & $10$ & m \\
Receiver depth & $z_r$ & $10$ & $10$ & m \\
Boundary treatment & --- & \multicolumn{2}{c}{Sponge layer (quadratic damping)} & --- \\
Absorbing-layer thickness & $n_b$ & $120$ & $120$ & grid points \\
\midrule
Grid footprint & --- & $700\times700$ & $700\times1900$ & m$^2$ \\
Seismic data shape & $(N_s,N_t,N_r)$ & $(5,1000,70)$ & $(5,1000,190)$ & --- \\
\bottomrule
\end{tabular}
\end{table}
\FloatBarrier

\subsection{Method Implementation Protocols}
\label{sec:exp_baselines}
\label{sec:appendix_baselines}
\label{sec:exp_conditions}
\label{sec:appendix_psg_implementation}

All methods use the same acoustic operator and observed waveforms, with
velocity coordinates normalized as in Appendix~\ref{sec:appendix_pretraining}.
Below, $x$ denotes a normalized velocity field and $\mathcal L_y(x)$ the
waveform loss evaluated after conversion to physical units.
The diffusion-based methods share the pretrained EDM for each family;
VP denoiser calls use the conversion in
Eq.~\eqref{eq:appendix_vp_ve_correspondence}.

\paragraph{FWI.}
Unregularized FWI minimizes $\mathcal L_y(x)$ by iteratively updating the
velocity using the adjoint waveform gradient~\cite{tarantola1984inversion,virieux2009overview}.
It uses the seismic observations alone to constrain the reconstruction.

\paragraph{Tikhonov-FWI.}
Tikhonov-FWI adds a first-order regularization term to the waveform
loss, penalizing squared differences between neighboring
velocities~\cite{asnaashari2013regularized},
\begin{equation}
  \label{eq:baseline_tikhonov}
  \Phi_{\mathrm{Tik}}(x)
    =\frac{1}{m}\sum_{i,j}\left[
      (x_{i+1,j}-x_{i,j})^2+(x_{i,j+1}-x_{i,j})^2
    \right],
\end{equation}
where $m=m_xm_z$ is the number of grid points and differences are taken
only between valid neighbors.
This quadratic penalty in Eq.~\eqref{eq:baseline_tikhonov}
suppresses spatial oscillations but can smooth
out sharp interfaces and fine-scale structures.

\paragraph{TV-FWI.}
TV-FWI adds a spatial total-variation regularizer~\cite{esser2018total}.
We use the anisotropic form adopted in the RED-DiffEq baseline
implementation~\cite{shan2026regularization},
\begin{equation}
  \label{eq:baseline_tv}
  \Phi_{\mathrm{TV}}(x)
    =\frac{1}{m}\sum_{i,j}\left[
      |x_{i+1,j}-x_{i,j}|+|x_{i,j+1}-x_{i,j}|
    \right],
\end{equation}
which favors piecewise-constant velocities while preserving sharp
contrasts.
The penalty in Eq.~\eqref{eq:baseline_tv} can introduce staircase patterns
where the true velocity varies smoothly.

\paragraph{DPS.}
Direct DPS conditions the reverse diffusion trajectory on the seismic
observations by evaluating the waveform loss at the denoised velocity
and differentiating through both the wave solver and the
denoiser~\cite{chung2022diffusion}.
The algorithmic procedure is discussed in detail in
Section~\ref{sec:method_dps}; our implementation uses residual-normalized
guidance (see Eq.~\eqref{eq:dps_guided_score}) for a stochastic VP proposal
and reports the final denoised velocity.

\paragraph{RED-DiffEq.}
RED-DiffEq augments the waveform gradient with the residual between
predicted and injected diffusion noise, thereby using the learned prior
as a stochastic regularizer.
For a uniformly sampled diffusion level $t$ and
$\epsilon\sim\N(0,I)$, its gradient-descent form is
\begin{equation}
  \label{eq:baseline_red_step}
  \begin{aligned}
    x_t^{(k)}
      &=\sqrt{\bar\alpha_t}\,x^{(k)}
        +\sqrt{1-\bar\alpha_t}\,\epsilon,\\
    x^{(k+1)}
      &=x^{(k)}-\eta_k\left[\nabla_x\mathcal L_y(x^{(k)})
        +\frac{\lambda}{m}
          \bigl(\widehat\epsilon_{\theta^\star}(x_t^{(k)},t)-\epsilon\bigr)\right].
  \end{aligned}
\end{equation}
Here $m=m_xm_z$ accounts for pixelwise mean reduction, and
$\widehat\epsilon_{\theta^\star}(x_t,t)=
(x_t-\sqrt{\bar\alpha_t}D_{\theta^\star}(x_t,t))/\sqrt{1-\bar\alpha_t}$
is the noise prediction obtained from the shared denoiser.
The implementation applies Adam to this direction in
Eq.~\eqref{eq:baseline_red_step}, treating the predicted
noise as fixed in the regularization term so that no denoiser Jacobian
is evaluated, following Algorithm~1 of Shan et al.~\cite{shan2026regularization}.

\paragraph{DiffusionFWI.}
DiffusionFWI inserts waveform-fitting optimization into reverse diffusion,
alternating a DDPM prior update with a block of physical velocity
updates~\cite{wang2023prior}.
We implement this alternating optimization with the shared EDM and
acoustic solver.  In our experiments, sampling starts at an intermediate
diffusion level from the smoothed velocity, and a fresh Adam optimizer
is used at each level.

\paragraph{SGDS-FWI.}
Split Gibbs Diffusion Sampling for FWI alternates likelihood-driven
physical refinement with diffusion denoising that encourages geological
structure~\cite{shen2026diffusion}.
Following the authors' released workflow, we first refine the initial
velocity by FWI--TV and SDEdit, then alternate warm-start FWI blocks,
decaying exploration noise, and one-step Tweedie denoising, using the
shared EDM and acoustic solver.

\paragraph{PSG.}
PSG uses persistent Adam for the physical state and an Euler--Maruyama prior
proposal for the reverse VP SDE~\cite{song2020score,zheng2025inversebench}.
Following DPS~\cite{chung2022diffusion}, feedback differentiates the Euclidean
norm of the residual between the denoised prediction and the updated physical
velocity, holding the latter fixed.  This scales the squared-residual gradient
by the inverse residual magnitude, with a squared-residual floor of
$10^{-20}$ to prevent division by zero.  After each physical Adam update,
we clip the normalized
velocity pointwise to $[-1,1]$, corresponding to $1500$--$4500$~m/s, while
retaining the optimizer state.  Values outside these physical bounds are
set to the nearest endpoint.  The terminal denoised estimate is also clipped
to these bounds before conversion to physical units.
At fixed positive diffusion time, the Euler and ancestral prior proposals
agree to first order in step size, while their finite-step means and
variances differ~\cite{ho2020denoising,song2020score}.

\paragraph{Domain decomposition for OOD inversion.}
To apply the pretrained prior to larger OOD fields, we adopt the domain
decomposition strategy of RED-DiffEq~\cite{shan2026regularization}, which
evaluates the prior on overlapping subdomains without retraining.
For the $70\times190$ OOD fields, the shared CurveFault-B prior acts on
three $70\times70$ patches with column ranges $[0,70)$, $[60,130)$,
and $[120,190)$.  Predictions are averaged with equal weights in each
ten-column overlap.  SGDS-FWI maintains independent patch states and
noise draws through each complete SDEdit or Tweedie call before folding
them into the full field.  The other learned methods use the same patch
positions and overlap averaging within their respective updates.
The OOD comparison comprises the three classical methods, RED-DiffEq,
DiffusionFWI, SGDS-FWI, and PSG.

\subsection{Hyperparameter Settings}
\label{sec:exp_hyperparameters}

PSG, the classical baselines, RED-DiffEq, and DiffusionFWI use the mean
$\ell_1$ waveform loss following the RED-DiffEq
repository~\cite{shan2026regularization}.
SGDS-FWI uses a squared-residual loss for physical optimization, following
the authors' implementation~\cite{shen2026diffusion}.
For DPS, we compute the squared-residual gradient and normalize it by the
residual magnitude to form the guidance, following the InverseBench
implementation of DPS~\cite{chung2022diffusion,zheng2025inversebench}.
Hyperparameters are fixed across the four OpenFWI families and all
observation conditions.
The classical methods and RED-DiffEq use $300$ Adam updates with
learning rate $0.03$ and cosine decay.
The regularization weights are $0.01$ for Tikhonov-FWI and TV-FWI and
$0.75$ for RED-DiffEq, following the empirical settings in
RED-DiffEq~\cite{shan2026regularization}.
The spatial penalties use directional edge averages, with a smoothed
absolute value for TV; RED-DiffEq uses no additional time weighting.

For DiffusionFWI, we use $10$ physical optimization steps between
consecutive denoising steps, following the inner-loop setting reported in
the original paper~\cite{wang2023prior}.
In our implementation, reverse diffusion starts at $t=100$ on the full
$T=1000$ DDPM schedule.  The physical updates use Adam at learning rate
$0.01$, with velocity-blur scale $0.2$ grid cells and gradient clipping
coefficient $1$.
DiffusionFWI shares the linear schedule in Algorithm~\ref{alg:vp_schedule}
with RED-DiffEq.
For SGDS-FWI, we retain the authors' sigmoid diffusion schedule, starting
level $250$, and exploration scale $0.3$~\cite{shen2026diffusion}.
We adapt the physical optimizer to Adam and use $300$ FWI--TV seed updates
followed by five blocks of $150$ updates at constant learning rate $0.03$.

DPS and PSG use the continuous VP schedule specified in
Appendix~\ref{sec:appendix_pretraining}, with $1000$ stochastic steps.
For DPS, the VP sampling configuration and residual-normalized waveform
guidance follow the InverseBench
implementation~\cite{chung2022diffusion,zheng2025inversebench}.
We set its guidance weight to $0.01$ based on an extensive
hyperparameter sweep.

For PSG, we set the normalized physical-state feedback weight to $0.3$.
Each diffusion step includes one physical Adam update at a constant
learning rate of $0.03$, with coupling weight $0.5$ on the pixelwise
mean squared velocity residual.
The per-step computational cost is analyzed in Appendix~\ref{sec:appendix_cost}.

\input{content/appendix_ood_parameters}

\subsection{Computational Cost}
\label{sec:appendix_cost}

PSG performs one physical update and one denoiser evaluation per reverse
step, reusing the denoised prediction for physical regularization and
diffusion guidance. Table~\ref{tab:psg_step_profile} reports the per-step
runtime breakdown for a CurveFault-B test model, using FP16 denoiser
computation and FP64 wave propagation. All stages are timed with GPU
synchronization, excluding diagnostic metrics and logging. On the same
model and device, a PSG step costs approximately $164\%$ of a standard
FWI iteration and $121\%$ of an iteration of the diffusion-based solver
RED-DiffEq.
Physical forward and backward operations account for $61.8\%$ of PSG's
runtime. With the same physical solver, PSG and FWI have comparable
observed peak device memory footprints of $4.40$~GiB and $4.03$~GiB,
respectively, including caches and the CUDA context.

The baseline implementations are adapted from the original papers and
available author-released code, with the settings detailed in
Appendices~\ref{sec:exp_baselines} and~\ref{sec:exp_hyperparameters}.
These settings entail different computational costs.
FWI and RED-DiffEq use $300$ optimization iterations, whereas DPS and
PSG use a full $1000$-step discretization of the VP noise schedule
associated with DDPM~\cite{ho2020denoising}, with the corresponding
sampling cost. The experiments therefore compare reconstruction quality
under the adopted settings, without matching the total computational
budget across methods. Adopting second-order FWI optimizers and
accelerated diffusion samplers in PSG, adjusting their relative
inner-loop update counts, and evaluating the trade-off between runtime
and reconstruction quality are directions for future work.

\input{tables/psg_step_profile}
\FloatBarrier

%% file: content/appendix_ood_parameters.tex
% Included within Appendix B.4; no new numbered subsection.
\paragraph{Salt-specific adjustments.}
OOD inversion reuses the CurveFault-B EDM through the domain
decomposition strategy of RED-DiffEq~\cite{shan2026regularization}.
The irregular Salt geometry motivated target-specific adjustment of
prior weights, physical-update budgets, and stabilization settings.
The learned-method parameters in Table~\ref{tab:ood_salt_parameters}
were selected using clean-target SSIM on Salt and then fixed across
its initialization and noise conditions; PSG uses two physical updates
per diffusion step.
The classical methods retain the OpenFWI learning rates and
regularization weights, with $1000$ updates.

\begin{table}[!htbp]
\centering\small
\caption{Selected Salt settings for the diffusion-based methods. Parameters
not listed retain the corresponding OpenFWI settings in
Appendix~\ref{sec:exp_hyperparameters}.}
\label{tab:ood_salt_parameters}
\begin{tabular}{lp{0.74\linewidth}}
\toprule
Method & Selected configuration \\
\midrule
RED-DiffEq & $1000$ updates; learning rate $0.1$ with cosine decay;
regularization $0.75$; sigmoid beta schedule; no time weighting. \\
DiffusionFWI & $100$ diffusion levels with $25$ FWI updates each;
learning rate $0.03$; horizontal/vertical gradient smoothing $0.5/1.0$;
velocity blur $0.4$; gradient normalization disabled; clipping coefficient $1$. \\
SGDS-FWI & $300$ seed FWI updates, five blocks of $1000$ updates,
and $100$ terminal updates; later-block and terminal TV weights $100$
with mean spatial reduction in physical velocity units; Tweedie
forward-noise scale $0.5$; terminal learning rate $0.01$. \\
PSG & $1000$ diffusion steps; feedback $0.8$; coupling $0.1$;
two physical updates per step; physical learning rate $0.03$ with cosine decay. \\
\bottomrule
\end{tabular}
\end{table}

\FloatBarrier

%% file: tables/psg_step_profile.tex
% Generated by profiling/build_psg_profile_table.py from synchronized measurements.
\begin{table}[!htbp]
\centering\small
\caption{Per-step runtime breakdown of PSG on a $70\times70$ OpenFWI
velocity model with five sources, measured on one NVIDIA GeForce RTX 5090
and averaged over $1000$ reverse steps after warm-up. $\pm$ denotes one
sample standard deviation. Diagnostic metrics and logging are excluded.}
\label{tab:psg_step_profile}
\begin{tabular}{lrr}
\toprule
Operation & Time (s) & Fraction \\
\midrule
EDM denoiser forward & $0.0362 \pm 0.0049$ & $25.3\%$ \\
Forward PDE solver and loss & $0.0305 \pm 0.0007$ & $21.3\%$ \\
Physical backward and Adam update & $0.0579 \pm 0.0011$ & $40.5\%$ \\
Denoiser backward and guidance & $0.0178 \pm 0.0023$ & $12.4\%$ \\
Reverse diffusion update & $0.0007 \pm 0.0003$ & $0.5\%$ \\
\midrule
Total per PSG step & $0.1431 \pm 0.0064$ & $100\%$ \\
\bottomrule
\end{tabular}
\end{table}